%% file: latex_template-main/template/sample.tex
\documentclass{applemlr}
\input{apple_preamble}

\usepackage{booktabs,tabularx,makecell,multirow,xcolor,pifont}

\usepackage{listings}
\usepackage{xcolor}

\definecolor{codegray}{gray}{0.45}
\definecolor{codeblue}{rgb}{0.2,0.2,0.7}
\definecolor{codegreen}{rgb}{0.0,0.5,0.0}

\usepackage{bbm}
\usepackage{xcolor}
\usepackage{colortbl}
\usepackage[most]{tcolorbox}

\usepackage{titletoc}

\usepackage[toc,page,header]{appendix}

\usepackage{algorithm}
\usepackage{bbm}
\usepackage[utf8]{inputenc} 
\usepackage[T1]{fontenc}    
\usepackage{url}            
\usepackage{booktabs}       
\usepackage{amsfonts}       
\usepackage{nicefrac}       
\usepackage{microtype}      
\usepackage{xcolor}         

\usepackage{pgfplots}
\pgfplotsset{compat=1.18}
\usetikzlibrary{patterns,patterns.meta}

\usepackage{pgfplots}
\pgfplotsset{compat=1.18}

\usepackage{multirow}
\usepackage{booktabs}
\usepackage{graphicx}
\usepackage[utf8]{inputenc} 
\usepackage[T1]{fontenc}    
\usepackage{url}            
\usepackage{amsfonts}       
\usepackage{nicefrac}       
\usepackage{microtype}      
\usepackage{xcolor}         
\usepackage{colortbl}
\usepackage{utfsym}
\usepackage{bbding}
\usepackage{booktabs}       

\usepackage{xcolor}
\usepackage{graphicx}
\usepackage{wrapfig}

\usepackage[most]{tcolorbox}
\usepackage{array}
\usepackage{caption}
\usepackage{enumitem}

\definecolor{deeporange}{rgb}{0.8, 0.3, 0.0}

\usepackage{tikz}
\usepackage{xcolor}
\usepackage{enumitem}

\usepackage{booktabs,tabularx,makecell,multirow,xcolor,pifont}

\usepackage{pifont}
\usepackage{amssymb}

\usepackage{pifont,xcolor}
\newcommand{\cmark}{\textcolor{green!50!black}{\ding{51}}} 
\newcommand{\xmark}{\textcolor{red}{\ding{55}}}

\usepackage{booktabs,tabularx,makecell,multirow,xcolor,pifont,array}
\newcolumntype{Y}{>{\centering\arraybackslash}X}

\usepackage{booktabs,tabularx,makecell,multirow,xcolor,pifont,array}
\newcolumntype{Y}{>{\centering\arraybackslash}X}

\definecolor{light-gray}{gray}{0.6}
\definecolor{front-color}{HTML}{F5FFFA}
\definecolor{Gray}{gray}{0.93}

\usepackage{times}
\usepackage{latexsym}
\usepackage{enumitem}
\usepackage{amsmath}
\usepackage{fixltx2e}
\usepackage{xspace}
\usepackage{stackengine}
\usepackage{amssymb}

\usepackage{tabu}
\usepackage{tikz}
\usepackage{mdframed}
\usepackage{tcolorbox}
\usepackage{cleveref}
\usepackage{microtype}
\usepackage{fontawesome}

\usepackage{tcolorbox}   
\usepackage{graphicx}    
\usepackage{array}       
\usepackage{colortbl}    
\usepackage{arydshln}    
\usepackage{xcolor}      

\definecolor{SummBlue}{RGB}{214,232,255}   
\definecolor{AssocGreen}{RGB}{215,242,220} 
\definecolor{NextOrange}{RGB}{255,234,210} 
\definecolor{ReIDYellow}{RGB}{255,246,210} 
\definecolor{TempPurple}{RGB}{236,225,255} 
\definecolor{NarrTeal}{RGB}{214,240,242}   

\definecolor{d8d9ed}{HTML}{d8d9ed}
\definecolor{f5f8ff}{HTML}{f5f8ff}
\colorlet{ThemeColor}{f5f8ff}

\newcommand*\colourcheck[1]{%
  \expandafter\newcommand\csname #1check\endcsname{\textcolor{#1}{\ding{52}}}%
}
\colourcheck{blue}
\colourcheck{green}
\colourcheck{red}

\usepackage{amssymb}
\usepackage{pifont}
\usepackage{xcolor}
\usepackage{booktabs}

\usepackage{float}
\usepackage{paralist}
\usepackage{multirow}
\usepackage{enumitem}
\usepackage{tikz}
\usepackage{pgfplots}
\pgfplotsset{compat=1.17}
\usepackage{graphicx}
\usepackage{stfloats}    
\usepackage{listings}
\usepackage{framed}
\usepackage{tcolorbox}

\usepackage[table]{xcolor}
\usepackage{booktabs}
\usepackage{multirow}
\usepackage{geometry}
\newcommand{\customsubsection}[1]{%
  \par
  \pagebreak[2]%
  \refstepcounter{subsection}%
    \everypar={%
      {\setbox0=\lastbox}
      \addcontentsline{toc}{subsection}{%
        {\protect\makebox[0.3in][r]{\thesubsubsection.} \hspace*{3pt}#1}}%
      \textbf{\thesubsection\space\space{#1}\space\newline}%
      \everypar={}%
    }%
  \ignorespaces
}

\def\ourbenchmark{\textsc{AMusE}\xspace} 
\def\oursolution{\texttt{RAFT}\xspace}

\definecolor{deepskyblue}{rgb}{0.0, 0.75, 1.0}
\definecolor{sweetpink}{rgb}{1.0, 0.8, 0.86}
\definecolor{rosepink}{rgb}{0.25, 0.55, 0.85}
\hypersetup{
    colorlinks,
    linkcolor={red},
    citecolor={rosepink}
}
\definecolor{iccvblue}{rgb}{0.21,0.49,0.74}

\def\ourbenchmark{\textsc{AMusE}\xspace} 
\def\oursolution{\texttt{RAFT}\xspace}

\definecolor{cvprblue}{rgb}{0.21,0.49,0.74}

\title{\raisebox{-0.1\height}{\includegraphics[width=0.06\linewidth]{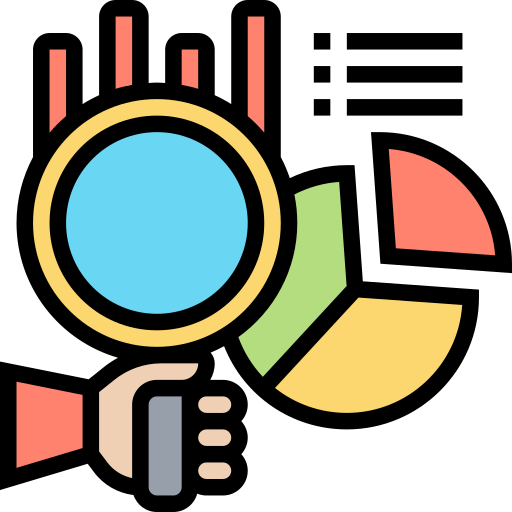}}\hspace{0.1em}\ourbenchmark: Audio-Visual Benchmark and Alignment Framework for Agentic Multi-Speaker Understanding}

\author{
Sanjoy Chowdhury$^{1,2,\dagger}$ \quad Karren D. Yang$^2$ \quad Xudong Liu$^2$ \quad Fartash Faghri$^2$ \quad Pavan Kumar Anasosalu Vasu$^2$ \quad Oncel Tuzel$^2$ \quad
Dinesh Manocha$^1$ \quad Chun-Liang Li$^2$ \quad Raviteja Vemulapalli$^2$ 
\vspace{1mm} \\
}

\affiliation{$^1$University of Maryland, College Park \quad $^2$Apple \quad}

\abstract{
Recent multimodal large language models (MLLMs) such as GPT-4o and Qwen3-Omni show strong perception but struggle in multi-speaker, dialogue-centric settings that demand agentic reasoning tracking who speaks, maintaining roles, and grounding events across time. These scenarios are central to multimodal audio-video understanding, where models must jointly reason over audio and visual streams in applications such as conversational video assistants and meeting analytics. We introduce \ourbenchmark, a benchmark designed around tasks that are inherently agentic, requiring models to decompose complex audio-visual interactions into planning, grounding, and reflection steps. It evaluates MLLMs across three modes zero-shot, guided, and agentic and six task families, including spatio-temporal speaker grounding and multimodal dialogue summarization. Across all modes, current models exhibit weak multi-speaker reasoning and inconsistent behavior under both non-agentic and agentic evaluation. Motivated by the inherently agentic nature of these tasks and recent advances in LLM agents, we propose \oursolution, a data-efficient agentic alignment framework that integrates reward optimization with intrinsic multimodal self-evaluation as reward and selective parameter adaptation for data and parameter efficient updates. Using \oursolution, we achieve up to $39.52\%$ relative improvement in accuracy on our benchmark. Together, \ourbenchmark and \oursolution provide a practical platform for examining agentic reasoning in multimodal models and improving their capabilities.}

\metadata[Correspondence]{\sffamily Sanjoy Chowdhury: \url{sanjoyc@umd.edu}; Raviteja Vemulapalli: \url{r_vemulapalli@apple.com}}
\date{\sffamily\today}

\begin{document}

\maketitle



\applefootnote{\textsuperscript{†} Work done during an internship at Apple.}
\section{Introduction}
\label{sec:intro}

Recent advances in multimodal large language models (MLLMs) such as \textit{GPT-4o}, and \textit{Qwen3-Omni} have achieved remarkable progress in visual understanding, instruction following, and cross-modal reasoning. As these systems evolve from passive perception to real-world \emph{agents}: meeting assistants, collaborative companions, or dialogue moderators they increasingly encounter multi-speaker, temporally grounded interactions. Understanding who speaks, maintaining role continuity, and reasoning coherently across participants are essential for socially consistent communication. Yet, such multi-speaker, role-aware understanding remains underexplored and largely unevaluated.

Existing multimodal benchmarks such as MMBench~\cite{liu2023mmbench}, MME~\cite{fu2023mme}, and MMMU~\cite{yue2024mmmu} primarily assess perception and single-turn reasoning, while dialogue-centric datasets such as M3Exam~\cite{zhang2023m3exam} and Video-ChatGPT~\cite{maaz2023video} focus on language quality without attributing reasoning to specific speakers. Even long-horizon evaluations like MMRC~\cite{xue2025mmrc} and MMLU-Pro~\cite{wang2024mmlu} assume a single narrator, overlooking inter-speaker transitions and shared context. Consequently, the ability of MLLMs to maintain speaker identity, resolve cross-turn dependencies, and perform structured reasoning in multi-speaker settings remains unknown.

\begin{figure*}[h]
\centering
\includegraphics[width=1\textwidth]{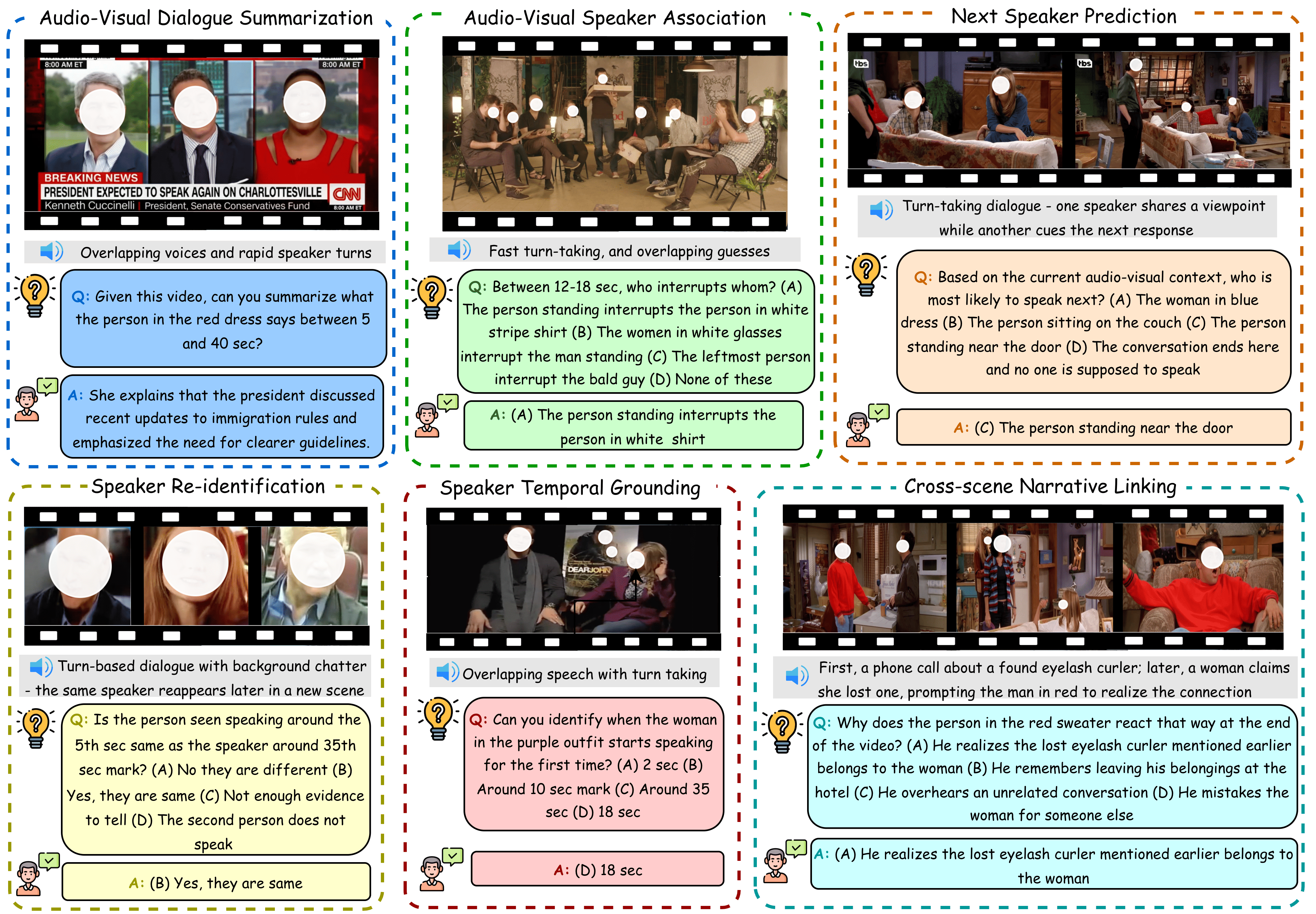}
\caption{\textbf{\ourbenchmark task definition.} The benchmark includes six high-level audio-visual tasks in realistic multi-speaker settings. Each task requires integrating core skills involve spatial and temporal grounding, speaker identification, speech recognition, and summarization. \includegraphics[height=0.9em]{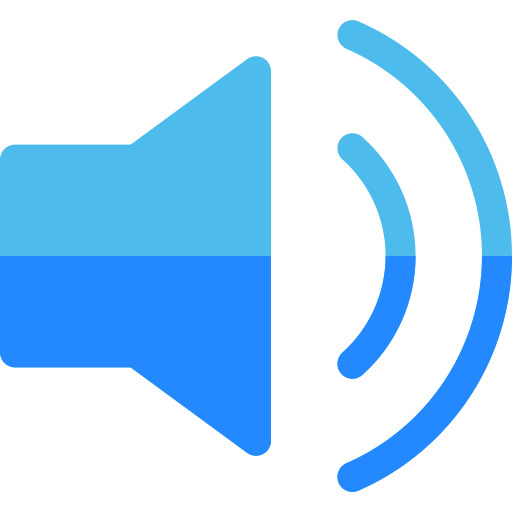} denotes a textual description of the audio scene for reader clarity.}
\label{fig:teaser}
\end{figure*}

Lately, the field is moving toward \emph{agentic multimodal systems} models capable not only of understanding but also of planning, acting, and reflecting~\cite{shinn2024reflexion,wang2024agentbench,liu2025agentboard}. However, existing evaluations of ``agency'' are limited to tool use or environment control for web agents \cite{ashraf2025agent}, T2I generation \cite{wang2025mllm} etc or text modality \cite{deng2025structuring} leaving open the question of how such autonomy translates to multi-party, audio-visual dialogue. Evaluating MLLMs in this setting requires decomposing high-level tasks into smaller perceptual and reasoning subgoals grounding, association, prediction, and summarization, thus demanding an explicitly \emph{agentic} mode of evaluation.

To address this gap, we present \textbf{\ourbenchmark}, a unified benchmark for assessing and improving multi-speaker agentic reasoning in MLLMs. 
It introduces three progressive modes of evaluation \emph{zero-shot}, \emph{guided}, and \emph{agentic} (Fig. \ref{fig:eval-modes}) spanning six tasks (Fig.\ref{fig:teaser}). 
These tasks are inherently agentic, requiring the model to plan, act, and reflect over multi-step interactions rather than rely on passive recognition. 
Our benchmark provides a structured testbed to analyze how autonomy, supervision, and reasoning granularity affect multimodal alignment and consistency.
This evaluation exposes persistent weaknesses of existing models in multi-speaker agentic reasoning, highlighting the need for alignment strategies tailored to this regime.

Motivated by the inherently agentic nature of these tasks, recent advances in LLM-based agents, and the gaps revealed by \ourbenchmark, we propose \textbf{\oursolution} (Reasoning–Acting–Feedback Training), a data-efficient alignment strategy that enhances agentic reasoning in MLLMs.
\oursolution integrates two key components: 
(i) \textit{Reflective Reward Optimization (RRO)} an intrinsic reward mechanism that reinforces multimodal and temporal correctness through self-evaluation, and 
(ii) \textit{Selective Reasoning Adaptation (SRA)} that focuses on updating cross-modal reasoning layers only for data efficiency and interpretability. 
Together, these components improve performance on \ourbenchmark by a margin and establish a practical framework for improving structured reasoning in MLLMs that future work can readily build upon.



\begin{table*}[t!]
\centering
\small
\resizebox{\linewidth}{!}{
\begin{tabular}{
c|
>{\centering\arraybackslash}m{1.4cm}|
>{\centering\arraybackslash}m{1.8cm}|
>{\centering\arraybackslash}m{1.6cm}|
>{\centering\arraybackslash}m{1.6cm}|
>{\centering\arraybackslash}m{1.8cm}|
ccc|ccc
}
\toprule
\multirow{2.5}{*}{\textbf{Benchmark}} &
\multirow{2.5}{*}{\textbf{\shortstack{Unique\\Spk}}} &
\multirow{2.5}{*}{\textbf{Avg Spk/Clip}} &
\multirow{2.5}{*}{\textbf{\shortstack{Overlap\\$\geq$2}}} &
\multirow{2.5}{*}{\textbf{\shortstack{Audio-Visual}}} &
\multirow{2.5}{*}{\textbf{\#Agentic Tasks}} &
\multicolumn{3}{c|}{\textbf{Reasoning Type}} &
\multicolumn{3}{c}{\textbf{Challenge Type}} \\
\cmidrule(lr){7-9} \cmidrule(lr){10-12}
& & & & & & Temporal & Causal & Identity & Grounding & Attribution & Summarization \\
\midrule
AVA-ActiveSpeaker~\cite{roth2020ava} &
430 & 2.5 & 0.12 &
\xmark & \xmark &
\xmark & \xmark & \xmark &
\cmark & \xmark & \xmark \\

VoxConverse~\cite{chung2020spot} &
800+ & 1.5 & 0.10 &
\xmark & \xmark &
\xmark & \xmark & \cmark &
\cmark & \cmark & \xmark \\

AMI Meeting Corpus~\cite{kraaij2005ami} &
30 & 4.0 & 0.22 &
\xmark & \xmark &
\cmark & \xmark & \cmark &
\cmark & \xmark & \xmark \\

Friends-MMC~\cite{wang2025friends} &
10 & 3.2 & 0.18 &
\xmark & \xmark &
\cmark & \cmark & \cmark &
\xmark & \cmark & \cmark \\

EgoSchema~\cite{egoschema2023} &
100+ & 1.8 & 0.12 &
\xmark & \xmark &
\cmark & \cmark & \xmark &
\xmark & \xmark & \xmark \\

Video-MME~\cite{fu2024video} &
150 & 2.1 & 0.15 &
\xmark & \xmark &
\cmark & \cmark & \xmark &
\xmark & \cmark & \cmark \\

\rowcolor{cyan!15}
\textbf{\ourbenchmark (Ours)} &
\textbf{350+} & \textbf{3.1} & \textbf{0.28} &
\cmark & \textbf{6} &
\cmark & \cmark & \cmark &
\cmark & \cmark & \cmark \\
\bottomrule
\end{tabular}}
\caption{\textbf{Comparison of multi-speaker audio-visual benchmarks.} \ourbenchmark uniquely integrates temporal, causal, and identity-based reasoning within overlapping multi-speaker settings. Unlike prior datasets focused on perception-only tasks, \ourbenchmark aligns audio-visual perception with structured reasoning to benchmark agentic, human-like understanding of multi-party discourse.}
\label{tab:amuse-comparison}
\end{table*}

Comprehensive experiments across both open- and closed-source MLLMs reveal two consistent findings: 
\textit{(a)} current MLLMs perform poorly across both non-agentic and agentic evaluations, underscoring the difficulty of sustained multi-speaker reasoning; and 
\textit{(b)} \oursolution substantially mitigates these gaps, yielding significant improvements in grounding accuracy, temporal consistency, and dialogue coherence. \textit{Our key contributions are}: \\
{\textbf{(1)}} We present \textcolor{blue}{\ourbenchmark, a comprehensive benchmark for evaluating multi-speaker understanding} by introducing \textit{six} challenging tasks. \\
{\textbf{(2)}} We perform a \textcolor{blue}{thorough evaluation under \textit{zero-shot}, \textit{guided}, and \textit{agentic} modes} of several open and closed source models. Our study reveals the limitations in the existing methods and provides a structured testbed for \textit{agentic} multi-modal assessment. \\
{\textbf{(3)}} We propose \textcolor{blue}{\oursolution, a model-agnostic and data-efficient alignment framework} integrating reward optimization (RRO) and data efficient update (SRA), yielding up to 39.52\% relative gain in accuracy.

\section{Related Works}

\noindent{\textbf{Multimodal Large Language Models.}}
The evolution of MLLMs has progressed from image-grounded models~\cite{liu2024visual,dai2023instructblip,zhu2023minigpt,zhang2023llavar,lai2024lisa,liu2024improved,wang2024cogvlm} to architectures capable of handling audio-visual dialogues and conversational contexts~\cite{shu2023audio,damonlpsg2025videollama3,wu2024next,li2025otter,wu2024deepseek,abouelenin2025phi,yan2024visa,zhang2024llava,zhang2025towards,guo2025seed1,wang2025skywork,zhu2025internvl3,comanici2025gemini}. This transition is driven by two key advances: (1) long-range context modeling for extended multi-turn dialogues~\cite{li2024videochatflash,song2024moviechat}, and (2) multimodal fusion of speaker identity, vocal traits, visual presence, and linguistic reasoning~\cite{ren2024timechat,maaz2024videogpt,yuan2025sa2va}. However, existing MLLMs remain constrained to simplified single-speaker or turn-level settings, lacking robust handling of overlapping speech, dynamic multi-speaker grounding, and cross-speaker temporal dependencies, limiting their applicability to real-world conversations such as meetings and panel discussions.

\noindent{\textbf{Video Benchmarks.}}
A comprehensive benchmark is crucial for evaluating MLLMs, yet existing efforts largely emphasize video understanding~\cite{lei2018tvqa,li2020hero,wu2024star,xu2017video,jang2017tgif,yu2019activitynet,mangalam2023egoschema} or general instruction-following~\cite{chen2024autoeval,liu2024tempcompass,fu2025video,li2024videovista,rawal2024cinepile,wang2024sok,ghermi2024short}. While these frameworks advance visual and temporal comprehension, they overlook challenges central to conversational audio-visual reasoning such as overlapping speech, multi-party grounding, speaker reidentification, and temporal coordination~\cite{ataallah2024infinibench,he2024mmworld,saravanan2025velociti}. Even long-context dialogue benchmarks~\cite{xiao2021next,li2024mvbench,ning2023video,xu2025beyond} often assume a single active speaker, limiting their capacity to assess multi-speaker role continuity. This absence of a dedicated benchmark hinders progress toward socially aware dialogue reasoning. To bridge this gap, we introduce \ourbenchmark, a unified framework for evaluating MLLMs in multi-speaker and agentic settings.


\noindent{\textbf{Agentic Systems.}} Recent progress in LMM-based agents has shown reasoning and planning abilities approaching human-level autonomy. Beyond single-agent systems~\cite{achiam2023gpt,touvron2023llama}, agentic frameworks leverage collective intelligence for complex tasks, from programming~\cite{li2023camel,dong2024self,hongmetagpt} to physical-world planning~\cite{dasgupta2022collaborating,song2023llm,huang2023inner,guo2024large}. Multi-agent setups further enhance generative capabilities~\cite{allioui2022multi,gal2024comfygen,wang2025genartist}, exemplified by DreamFactory~\cite{xie2024dreamfactory} for video synthesis, Mora~\cite{yuan2024mora} for human-guided refinement, and SPAgent~\cite{tu2024spagent} for tool orchestration. Unlike prior work, our approach introduces enhanced collaborative mechanisms for executing cross-modal reasoning tasks for speaker-specific settings.

\section{\ourbenchmark: \underline{A}gentic \underline{Mu}lti-\underline{s}peaker Understanding \& \underline{E}valuation}


\begin{table*}[t!]
\begin{minipage}[b]{0.78\textwidth}
\centering
\includegraphics[width=1\textwidth]{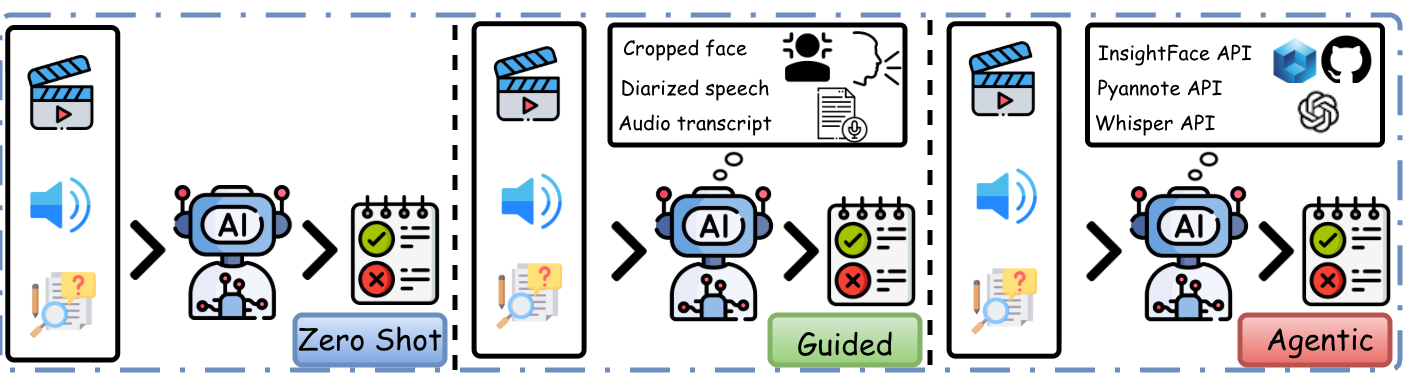}
\captionof{figure}{\textbf{Evaluation Protocols.} \textit{Zero-Shot}, \textit{Guided}, and \textit{Agentic} where MLLMs reason over raw input, use auxiliary cues (e.g., faces, transcripts), or invoke external tools (e.g., Whisper, Pyannote, InsightFace).}
\label{fig:eval-modes}
\end{minipage}
\hfill
\begin{minipage}[b]{0.20\textwidth}
\centering
\scriptsize
\renewcommand\arraystretch{1.4}
\setlength{\tabcolsep}{3mm}
\resizebox{\linewidth}{!}{
\begin{tabular}{c|cccc}
\toprule
\multirow{2}{*}{\vspace{0.2em}\textbf{Task}} & \multicolumn{4}{c}{\textbf{Additional Metadata}} \\
 & \textbf{FC} & \textbf{VS} & \textbf{TR} & \textbf{LS} \\
\midrule
STG  & {\color{green!60!black}\checkmark} & {\color{red!70!black}\xmark} & {\color{red!70!black}\xmark} & {\color{green!60!black}\checkmark} \\
AVDS & {\color{green!60!black}\checkmark} & {\color{green!60!black}\checkmark} & {\color{green!60!black}\checkmark} & {\color{red!60!black}\checkmark} \\
AVSA & {\color{green!60!black}\checkmark} & {\color{green!60!black}\checkmark} & {\color{red!70!black}\xmark} & {\color{red!70!black}\xmark} \\
NSP  & {\color{red!70!black}\xmark} & {\color{red!70!black}\xmark} & {\color{green!60!black}\checkmark} & {\color{green!60!black}\checkmark} \\
SRID & {\color{green!60!black}\checkmark} & {\color{red!70!black}\xmark} & {\color{red!70!black}\xmark} & {\color{green!60!black}\checkmark} \\
CSNL & {\color{green!60!black}\checkmark} & {\color{green!60!black}\checkmark} & {\color{green!60!black}\checkmark} & {\color{green!60!black}\checkmark} \\
\bottomrule
\end{tabular}}
\captionof{table}{\textbf{Task-wise use of additional cues}: Face Crops (FC), Voice Segments (VS), Transcripts (TS), and Lip Sync (LS).}
\label{tab:perceptual_tools}
\end{minipage}
\end{table*}

\customsubsection{Task Definition}



Our benchmark systematically evaluates MLLMs in complex, multi-speaker, dialogue-rich environments. Unlike prior multimodal benchmarks that focus on perception or single-turn reasoning, \ourbenchmark targets multi-speaker reasoning, requiring models to identify speakers, maintain temporal and social context, and reason across conversational turns and scenes. We define \textit{six} representative tasks (refer to Fig. \ref{fig:teaser}) to capture these dimensions.

\noindent\textbf{Audio-Visual Dialogue Summarization (AVDS).}
AVDS evaluates a model’s ability to produce coherent summaries of multi-person conversations while retaining the correct speaker roles and content attribution. This task is crucial for assessing high-level comprehension, as summarization reflects whether a model can integrate linguistic, visual, and temporal cues into structured discourse understanding. Its difficulty arises from dense speaker interactions, overlapping utterances, and role entanglement requiring models to attend selectively across modalities while preserving conversational flow.

\noindent\textbf{Audio-Visual Speaker Association (AVSA).}
AVSA maps each spoken utterance to its corresponding visible speaker an essential step for identity-aware dialogue understanding and higher-level reasoning such as attribution and summarization. The task is challenging as it requires fine-grained \textit{cross-modal disambiguation} in natural conversations. For example, in a group game setting (Fig.\ref{fig:teaser}), several participants may talk or laugh simultaneously while one person interrupts another mid-sentence, requiring the model to detect who is actively speaking versus reacting. Success demands precise integration of phonetic cues, lip motion, gaze, and temporal interaction patterns to reliably associate voices with the correct individuals amid overlaps or occlusions.

\noindent\textbf{Next Speaker Prediction (NSP).} This task anticipates the next active speaker in a conversation, reflecting a model’s ability to capture social and turn-taking dynamics. 
It is important for modeling interactive conversational flow and understanding response patterns across individuals. 
The difficulty stems from the need for long-term temporal reasoning and social inference models must interpret subtle multimodal cues such as gaze shifts, pauses, and prosody to predict interaction intent.

\noindent\textbf{Speaker Re-identification (SRID).}
SRID evaluates a model’s ability to recognize and match speakers across non-contiguous clips essential for long-context tracking and dialogue continuity. The task is challenging since identity cues vary with appearance, viewpoint, lighting, background, or emotion, requiring modality-invariant embeddings that maintain consistent speaker representation under such variations.

\noindent\textbf{Speaker Temporal Grounding (STG).}
This task requires locating the time span and identity of the active speaker within a scene. It is important because temporal alignment between speech and visual context forms the foundation of time-sensitive multimodal understanding, enabling models to localize when and who is speaking. The task is challenging as it demands synchronizing acoustic cues, lip motion, and facial dynamics under overlapping or intermittent speech conditions, where multiple people may talk simultaneously and attention must dynamically shift across them.

\noindent\textbf{Cross-Scene Narrative Linking (CSNL).}
CSNL tests if models can connect events or utterances across distinct scenes to infer causal and temporal relationships. 
It is one of the most cognitively demanding tasks, as it requires reasoning beyond perceptual continuity to form coherent narratives spanning multiple contexts. 
The challenge lies in bridging long-range temporal gaps and integrating visual, auditory, and linguistic evidence to establish logical dependencies and narrative coherence across time.


\begin{figure*}[h]
\centering
\includegraphics[width=1\textwidth]{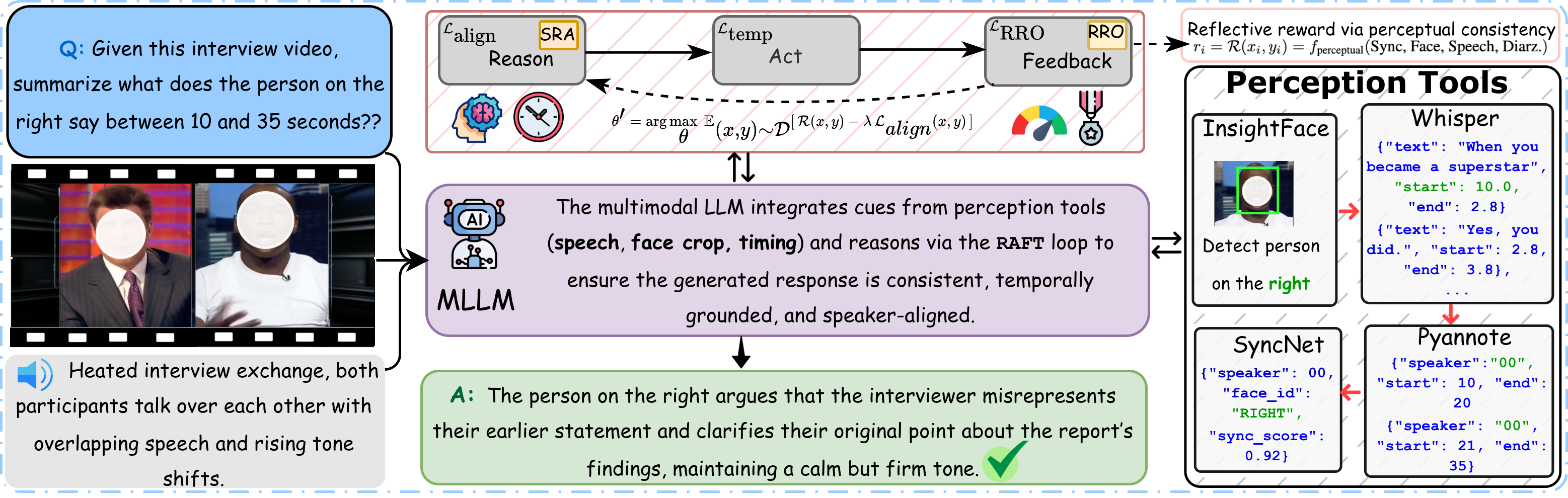}
\caption{\textbf{\oursolution framework for agentic multimodal reasoning.}
Given a dialogue-rich video, the model uses perception tools to extract multimodal cues. 
\oursolution integrates SRA and RRO within a Reason–Act–Feedback loop, using perceptual consistency to refine temporal and speaker-grounded responses. \oursolution(\includegraphics[height=0.9em]{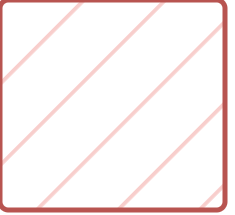}) module operates only during training. Dotted arrow shows that RRO passively uses perceptual feedback for reward computation rather than active control of the tools.}
\label{fig:raft}
\end{figure*}

\customsubsection{Evaluation Protocols}
\noindent\textbf{Zero-Shot Evaluation.}
\label{sec:zs-eval}
In this setting, each model is evaluated without external guidance or perceptual cues (Fig.~\ref{fig:eval-modes}). It receives only the raw video and the question as input. Zero-shot evaluation serves as the lower bound, reflecting the model’s intrinsic multimodal understanding from pretraining.


\noindent\textbf{Guided Evaluation.}
In this mode, the model receives auxiliary perceptual cues such as pre-computed face crops, active-speaker timestamps, or ASR transcripts, along with explicit guidance on their use (Tab. \ref{tab:perceptual_tools}).
Prompts include step-by-step instructions (e.g., ``Use the transcript to identify who is speaking before summarizing the dialogue'').
This setting tests how well models \emph{integrate structured cues} and follow task decomposition in a controlled setup, simulating an \emph{assisted agent} with external signals but limited autonomy.

\noindent\textbf{Agentic Evaluation.}
The most challenging regime removes all explicit hints about tool availability or intermediate steps. External modules (face detection, speaker localization, speech-to-text) remain accessible at runtime, but the model must discover and invoke them implicitly through its own reasoning (Fig. \ref{fig:eval-modes}). This setting mirrors a fully autonomous agent operating in open environments, inferring actions directly from goals. Performance here reflects true agentic understanding of how well the model self-directs perception, reasoning, and reflection across multi-speaker interactions without human scaffolding.

\begin{table*}[t!]
\centering
\tiny
\renewcommand\arraystretch{0.8}
\setlength{\tabcolsep}{1pt}
\resizebox{\linewidth}{!}{
\begin{tabular}{lcccccccccccccccc}
\toprule
\multirow{2}{*}{Model} & \multicolumn{4}{c}{Zero-Shot} & \multicolumn{4}{c}{Guided} & \multicolumn{4}{c}{Agentic} & \multicolumn{4}{c}{Agentic w/ \oursolution} \\
\cmidrule(lr){2-5} \cmidrule(lr){6-9} \cmidrule(lr){10-13} \cmidrule(lr){14-17}
 & BLEU$\uparrow$ & METEOR$\uparrow$ & CIDEr$\uparrow$ & GPT$\uparrow$ & BLEU$\uparrow$ & METEOR$\uparrow$ & CIDEr$\uparrow$ & GPT$\uparrow$ & BLEU$\uparrow$ & METEOR$\uparrow$ & CIDEr$\uparrow$ & GPT$\uparrow$ & BLEU$\uparrow$ & METEOR$\uparrow$ & CIDEr$\uparrow$ & GPT$\uparrow$ \\
\midrule
Human & 86.04 & 8.36 & 92.03 & 9.52 & -- & -- & -- & -- & -- & -- & -- & -- & -- & -- & -- & -- \\
Random & 21.88 & 2.14 & 30.74 & 2.23 & -- & -- & -- & -- & -- & -- & -- & -- & -- & -- & -- & -- \\
\midrule
\rowcolor{gray!20}
\multicolumn{17}{c}{\textbf{\textit{Closed-source MLLMs}}} \\
REKA & 39.14 & 3.19 & 38.72 & 3.27 & 41.03 & 4.46 & 42.55 & 4.56 & 38.07 & 3.21 & 30.03 & 3.05 & -- & -- & -- & -- \\
GPT-4o Mini & 42.13 & 3.87 & 40.36 & -- & 45.26 & 4.47 & 42.54 & -- & 43.17 & 4.59 & 35.11 & -- & -- & -- & -- & -- \\
GPT-4o & 43.52 & 4.99 & \cellcolor{cyan!15}\textbf{42.89} & -- & \cellcolor{cyan!15}\textbf{49.21} & \cellcolor{cyan!15}\textbf{5.36} & 50.19 & -- & 44.41 & \cellcolor{cyan!15}\textbf{4.98} & 44.01 & -- & -- & -- & -- & -- \\
\midrule
\rowcolor{gray!20}
\multicolumn{17}{c}{\textbf{\textit{Open-source MLLMs}}} \\
Unified-IO2-5B & 35.02 & 3.67 & 39.05 & 3.90 & 39.02 & 4.03 & 41.53 & 3.34 & 35.08 & 3.74 & 38.09 & 3.02 & \cellcolor[HTML]{F2FBFD}42.43 & \cellcolor[HTML]{F2FBFD}4.68 & \cellcolor[HTML]{F2FBFD}43.19 & \cellcolor[HTML]{F2FBFD}4.11 \\
CREMA & 36.04 & 3.68 & 37.00 & 4.34 & 40.50 & 4.39 & 44.08 & 3.76 & 34.04 & 3.67 & 40.08 & 3.33 & \cellcolor[HTML]{F2FBFD}43.18 & \cellcolor[HTML]{F2FBFD}4.36 & \cellcolor[HTML]{F2FBFD}46.85 & \cellcolor[HTML]{F2FBFD}4.46 \\
Video-SALMONN & 37.35 & 3.95 & 36.06 & 4.87 & 41.04 & 4.34 & 46.26 & 4.15 & 35.08 & 4.55 & 42.04 & 3.97 & \cellcolor[HTML]{F2FBFD}43.25 & \cellcolor[HTML]{F2FBFD}5.17 & \cellcolor[HTML]{F2FBFD}48.09 & \cellcolor[HTML]{F2FBFD}4.92 \\
VITA-8B & 41.02 & 3.70 & 40.05 & 4.94 & 42.57 & 4.29 & 47.05 & 4.34 & 38.54 & 4.85 & 42.56 & 4.07 & \cellcolor[HTML]{F2FBFD}49.58 & \cellcolor[HTML]{F2FBFD}5.88 & \cellcolor[HTML]{F2FBFD}48.09 & \cellcolor[HTML]{F2FBFD}5.10 \\
Qwen2.5-Omni-7B & 44.92 & 4.46 & 41.78 & 5.09 & 46.06 & 4.45 & 48.56 & 5.14 & 42.07 & 4.58 & 45.09 & 4.83 & \cellcolor[HTML]{F2FBFD}51.54 & \cellcolor[HTML]{F2FBFD}6.04 & \cellcolor[HTML]{F2FBFD}54.05 & \cellcolor[HTML]{F2FBFD}6.16 \\
Qwen3-Omni-7B & \cellcolor{cyan!15}\textbf{45.08} & \cellcolor{cyan!15}\textbf{5.26} & 42.03 & \cellcolor{cyan!15}\textbf{5.24} & 48.08 & 4.89 & \cellcolor{cyan!15}\textbf{50.62} & \cellcolor{cyan!15}\textbf{5.45} & \cellcolor{cyan!15}\textbf{45.07} & 4.72 & \cellcolor{cyan!15}\textbf{48.53} & \cellcolor{cyan!15}\textbf{5.10} & \cellcolor{cyan!15}\textbf{54.54} & \cellcolor{cyan!15}\textbf{6.81} & \cellcolor{cyan!15}\textbf{58.51} & \cellcolor{cyan!15}\textbf{6.62} \\
\bottomrule
\end{tabular}}
\caption{
\textbf{Audio-Visual Dialogue Summarization results.} While closed-source models such as GPT-4o achieve strong zero-shot and guided performance, open-source MLLMs benefit substantially from \oursolution training.}
\label{tab:avds}
\end{table*}

\begin{table*}[t!]
\centering
\tiny
\renewcommand\arraystretch{0.8}
\setlength{\tabcolsep}{1mm}
\resizebox{\linewidth}{!}{
\begin{tabular}{lcccccccccccc}
\toprule
\multirow{2}{*}{\textbf{Model}} &
\multicolumn{4}{c}{\textbf{AV Speaker Association (Acc.\% $\uparrow$)}} &
\multicolumn{4}{c}{\textbf{Next Speaker Prediction (Acc.\% $\uparrow$)}} &
\multicolumn{4}{c}{\textbf{Speaker Re-identification (Acc.\% $\uparrow$)}} \\
\cmidrule(lr){2-5} \cmidrule(lr){6-9} \cmidrule(lr){10-13}
 & ZS & Guided & Agentic & Agt w/R &
 ZS & Guided & Agentic & Agt w/R &
 ZS & Guided & Agentic & Agt w/R \\
\midrule
Human & 91.02 & – & – & – & 95.03 & – & – & – & 89.06 & – & – & – \\
Random & 25.40 & – & – & – & 25.64 & – & – & – & 25.79 & – & – & – \\
\midrule
\rowcolor{gray!20}\multicolumn{13}{c}{\textbf{\textit{Closed-source MLLMs}}}\\
REKA & 42.76 & 45.80 & 41.54 & -- & 41.89 & 44.52 & 41.50 & -- & 52.50 & 54.60 & 50.70 & -- \\
GPT-4o Mini & 44.27 & 46.18 & 43.78 & -- & 44.13 & 46.24 & 44.10 & -- & 53.27 & 56.97 & 53.33 & -- \\
GPT-4o & 47.18 & 48.98 & \cellcolor{cyan!15}\textbf{47.00} & -- &
\cellcolor{cyan!15}\textbf{49.18} & \cellcolor{cyan!15}\textbf{50.24} & \cellcolor{cyan!15}\textbf{46.58} & -- &
54.63 & 55.82 & 52.50 & -- \\
\midrule
\rowcolor{gray!20}\multicolumn{13}{c}{\textbf{\textit{Open-source MLLMs}}}\\
Unified-IO2-5B & 34.68 & 37.92 & 33.83 & \cellcolor[HTML]{F2FBFD}43.12 & 35.09 & 38.46 & 35.78 & \cellcolor[HTML]{F2FBFD}43.38 & 43.78 & 46.82 & 41.52 & \cellcolor[HTML]{F2FBFD}51.08 \\
CREMA & 37.33 & 41.04 & 37.06 & \cellcolor[HTML]{F2FBFD}45.81 & 40.09 & 43.57 & 40.56 & \cellcolor[HTML]{F2FBFD}44.09 & 47.34 & 50.36 & 47.88 & \cellcolor[HTML]{F2FBFD}52.73 \\
Video-SALMONN & 38.57 & 41.08 & 38.53 & \cellcolor[HTML]{F2FBFD}48.18 & 44.05 & 47.78 & 43.82 & \cellcolor[HTML]{F2FBFD}52.89 & 48.59 & 51.34 & 48.46 & \cellcolor[HTML]{F2FBFD}53.25 \\
VITA-8B & 40.74 & 42.51 & 40.01 & \cellcolor[HTML]{F2FBFD}43.72 & 44.57 & 42.59 & 41.60 & \cellcolor[HTML]{F2FBFD}48.47 & 50.76 & 52.50 & 50.58 & \cellcolor[HTML]{F2FBFD}52.98 \\
Qwen2.5-Omni-7B & 45.65 & 48.89 & 44.46 & \cellcolor[HTML]{F2FBFD}48.89 & 50.81 & 48.11 & 43.54 & \cellcolor[HTML]{F2FBFD}52.33 & 54.67 & 56.82 & 52.19 & \cellcolor[HTML]{F2FBFD}58.62 \\
Qwen3-Omni-7B & \cellcolor{cyan!15}\textbf{47.74} & \cellcolor{cyan!15}\textbf{49.55} & 46.98 & \cellcolor{cyan!15}\textbf{54.22} & 48.38 & 49.27 & 45.02 & \cellcolor{cyan!15}\textbf{56.73} & \cellcolor{cyan!15}\textbf{56.98} & \cellcolor{cyan!15}\textbf{58.65} & \cellcolor{cyan!15}\textbf{54.51} & \cellcolor{cyan!15}\textbf{62.53} \\
\bottomrule
\end{tabular}}
\caption{\textbf{Performance comparison on AV Speaker Association, Next Speaker Prediction, and Speaker Re-identification tasks.} Consistent performance gains for Qwen-based models after \oursolution fine-tuning. \textbf{Agt w/R}: Agentic evaluation with \oursolution finetuning.}
\label{tab:avsa-nsp-sri}
\end{table*}

\customsubsection{Sample Curation}
The dataset is constructed by curating examples from AVA Active Speaker~\cite{gu2018ava}, VoxCeleb2~\cite{chung2018voxceleb2}, FriendsMMC~\cite{wang2025friends}, AMI Meetings~\cite{kraaij2005ami}, and web-scraped sources. We focus on multi-person conversations such as talk shows, interviews, and podcasts etc. Each 10–50s clip contains $2-10$ visually and audibly active speakers. A semi-automated construction pipeline combines metadata and rule-based matching (more details in supplementary), followed by manual inspection for annotation sanity and consistency. The resulting dataset spans diverse acoustic conditions, speaking styles, and scene dynamics, capturing realistic conversational complexity.

We employ diverse templates to form task questions, ensuring linguistic variation. QA pairs evaluate temporal grounding, audio-visual association, and related reasoning aspects. Further dataset details are provided in the supplementary material.




\vspace{1mm}
\customsubsection{Benchmark Statistics}
\ourbenchmark contains $2,100$ samples spanning six tasks: STG, AVDS, AVSA, NSP, and SRID ($400$ each), and CSNL ($100$ manually collected cross-scene instances). It features one of the highest average speaker counts per clip among existing benchmarks (Tab.~\ref{tab:amuse-comparison}), with an average clip length of $38.7 s$, covering over $23$ hours of annotated multi-speaker content and $350+$ unique identities. This balanced structure ensures diversity across modalities and reasoning levels, supporting both low-level perceptual evaluation and high-level discourse reasoning.

\vspace{2mm}
\section{Our Approach: \underline{R}easoning, \underline{A}cting, and \underline{F}eedback \underline{T}raining (\oursolution)}

Motivated by the inherently agentic nature of our multi-speaker audio-visual tasks and advances in LLM agents,
we introduce \oursolution, a self-reflective alignment framework for multi-speaker audio-visual reasoning to allow future research to build upon.
\oursolution couples \emph{Reflective Reward Optimization} for reward-driven generative correction with \emph{Selective Reasoning Adaptation} for data and parameter efficient updates. 
Together, these modules enable fine-grained temporal grounding, adaptive speaker attribution, and stable low-resource alignment.

\customsubsection{Problem Setup}
\noindent
\begin{tcolorbox}[
    colback=blue!4!white,
    colframe=blue!35!black,
    boxsep=1pt,
    left=2pt,
    right=2pt,
    top=2pt,
    bottom=2pt,
    arc=1pt,
    boxrule=0.3pt
]
\small
\textbf{Input:} Multimodal streams $x = \{x^{(a)}, x^{(v)}, x^{(t)}\}$ representing audio, video, and text. 
\textbf{Output:} Structured response $y = \{p, a, r\}$ denoting the \textit{Plan}, \textit{Act}, and \textit{Reflect} phases.
\end{tcolorbox}


\noindent Given an input context $x$ composed of multimodal streams $\{x^{(a)}, x^{(v)}, x^{(t)}\}$  and a desired structured response $y = \{p, a, r\}$, we model a conditional policy $\pi_\theta(y|x)$ and seeks parameters $\theta'$ that maximize both reward and alignment:
\begin{equation}\footnotesize
    \theta' = \arg\max_\theta \; \mathbb{E}_{(x,y)\sim\mathcal{D}}[\,\mathcal{R}(x,y) - \lambda\,\mathcal{L}_{align}(x,y)\,],
\end{equation}
where $\mathcal{R}$ measures agentic correctness and $\mathcal{L}_{align}$ regularizes stepwise reasoning.

\customsubsection{Structured Reasoning Alignment}
\noindent
\begin{tcolorbox}[
    colback=blue!4!white,
    colframe=blue!35!black,
    boxsep=1pt,
    left=2pt,
    right=2pt,
    top=2pt,
    bottom=2pt,
    arc=1pt,
    boxrule=0.3pt
]
\small
\textbf{Input:} Reasoning phases $\{p, a, r\}$ and their contextual dependencies $\mathcal{C}_k$. 
\textbf{Output:} Temporally and semantically aligned reasoning segments.
\end{tcolorbox}

\noindent For each step $k$, the model predicts a target segment 
$y=k^*$ conditioned on the input $x$, where $x$ denotes history of observed history, and the model additionally conditions on the reasoning it has generated so far, which we omit here for notation simplicity.

\noindent The structured alignment objective is then written purely in terms of 
inputs $x$ and labels $y$:
\begin{equation}
\mathcal{L}_{\text{align}}(x, y)
=
-\log \pi_\theta(y \mid x)
\end{equation}

\noindent This encourages each phase to remain consistent with its contextual 
dependencies, enforcing coherent Plan-Act-Reflect reasoning.

\begin{table*}[t!]
\centering
\scriptsize
\renewcommand\arraystretch{0.9}
\setlength{\tabcolsep}{0.9mm}
\resizebox{\linewidth}{!}{
\begin{tabular}{lcccccccccccccccc}
\toprule
 & \multicolumn{8}{c}{\textbf{Speaker Temporal Grounding}} & \multicolumn{8}{c}{\textbf{Cross-Modal Narrative Linking}} \\
\midrule
\multirow{2}{*}{\textbf{Model}} &
\multicolumn{4}{c}{\textbf{Temporal IoU ($\uparrow$)}} &
\multicolumn{4}{c}{\textbf{Off-by-One Accuracy ($\uparrow$)}} &
\multicolumn{4}{c}{\textbf{Accuracy ($\uparrow$)}} &
\multicolumn{4}{c}{\textbf{Human-Judged Coherence ($\uparrow$)}} \\
\cmidrule(lr){2-5} \cmidrule(lr){6-9} \cmidrule(lr){10-13} \cmidrule(lr){14-17}
 & ZS & Guided & Agentic & Ag w/R & ZS & Guided & Agentic & Ag w/R & ZS & Guided & Agentic & Ag w/R & ZS & Guided & Agentic & Ag w/R \\
\midrule
Human  & 94.66 & – & – & – & 83.31 & – & – & – & 88.23 & – & – & – & -- & – & – & – \\
Random & 30.87 & – & – & – & 29.47 & – & – & – & 21.18 & – & – & – & 2.56 & – & – & – \\
\midrule
\rowcolor{gray!20}\multicolumn{17}{c}{\textbf{\textit{Closed-source MLLMs}}}\\
REKA & 43.52 & 50.31 & 46.08 & -- & 46.75 & 50.54 & 45.59 & -- & 40.20 & 46.33 & 42.78 & -- & 4.40 & 4.65 & 4.21 & -- \\
GPT-4o Mini & 44.21 & 49.41 & 45.98 & -- & 46.28 & 49.11 & 46.92 & -- & 41.89 & 47.29 & 45.29 & -- & 5.01 & 4.94 & 4.15 & -- \\
GPT-4o & \cellcolor{cyan!15}\textbf{48.72} & \cellcolor{cyan!15}\textbf{52.03} & \cellcolor{cyan!15}\textbf{49.17} & -- & 49.28 & \cellcolor{cyan!15}\textbf{52.11} & \cellcolor{cyan!15}\textbf{47.44} & -- &
43.16 & 49.26 & \cellcolor{cyan!15}\textbf{43.23} & -- &
\cellcolor{cyan!15}\textbf{5.22} & 6.02 & \cellcolor{cyan!15}\textbf{5.63} & -- \\
\midrule
\rowcolor{gray!20}\multicolumn{17}{c}{\textbf{\textit{Open-source MLLMs}}}\\
Unified-IO2-5B & 33.08 & 36.77 & 32.89 & \cellcolor[HTML]{F2FBFD}40.30 & 36.26 & 39.38 & 35.57 & \cellcolor[HTML]{F2FBFD}40.57 & 35.89 & 38.20 & 34.98 & \cellcolor[HTML]{F2FBFD}42.16 & 4.18 & 4.47 & 4.41 & \cellcolor[HTML]{F2FBFD}5.29 \\
CREMA & 34.26 & 35.58 & 28.93 & \cellcolor[HTML]{F2FBFD}37.27 & 38.65 & 41.53 & 32.57 & \cellcolor[HTML]{F2FBFD}41.04 & 39.48 & 39.59 & 33.11 & \cellcolor[HTML]{F2FBFD}42.82 & 4.93 & 4.75 & 4.38 & \cellcolor[HTML]{F2FBFD}5.42 \\
Video-SALMONN & 38.01 & 41.44 & 38.68 & \cellcolor[HTML]{F2FBFD}39.57 & 41.90 & 43.45 & 37.57 & \cellcolor[HTML]{F2FBFD}44.18 & 42.90 & 43.01 & 41.27 & \cellcolor[HTML]{F2FBFD}46.23 & 4.13 & 4.78 & 4.49 & \cellcolor[HTML]{F2FBFD}5.68 \\
VITA-8B & 41.18 & 46.27 & 42.29 & \cellcolor[HTML]{F2FBFD}49.22 & 44.47 & 48.67 & 41.11 & \cellcolor[HTML]{F2FBFD}50.48 & 43.27 & 44.83 & 41.26 & \cellcolor[HTML]{F2FBFD}45.50 & 4.56 & 4.60 & 4.05 & \cellcolor[HTML]{F2FBFD}4.81 \\
Qwen2.5-Omni-7B & 47.76 & 50.57 & 42.53 & \cellcolor[HTML]{F2FBFD}52.08 & 47.38 & 48.65 & 42.80 & \cellcolor[HTML]{F2FBFD}54.55 & 45.30 & 48.90 & 40.50 & \cellcolor[HTML]{F2FBFD}53.80 & 5.32 & 5.62 & 5.24 & \cellcolor[HTML]{F2FBFD}6.28 \\
Qwen3-Omni-7B & 48.56 & 51.02 & 45.59 & \cellcolor{cyan!15}\textbf{54.04} & \cellcolor{cyan!15}\textbf{49.51} & 51.50 & 43.29 & \cellcolor{cyan!15}\textbf{56.33} &
\cellcolor{cyan!15}\textbf{46.07} & \cellcolor{cyan!15}\textbf{49.76} & 41.04 & \cellcolor{cyan!15}\textbf{57.26} & 5.66 & \cellcolor{cyan!15}\textbf{6.27} & 5.02 & \cellcolor{cyan!15}\textbf{7.11} \\
\bottomrule
\end{tabular}}
\caption{\textbf{Performance on Speaker Temporal Grounding and Cross-scene Narrative Linking tasks.} Human-Judged Coherence is scaled between 0-10. 
\oursolution yields substantial gains in temporal precision and narrative coherence, especially for open-source MLLMs such as Qwen3-Omni. \textbf{Ag w/R}: Agentic evaluation with \oursolution finetuning.}
\label{tab:stg-csnl}
\end{table*}


\customsubsection{Reflective Reward Optimization (RRO)}
\noindent
\begin{tcolorbox}[
    colback=blue!4!white,
    colframe=blue!35!black,
    boxsep=1pt,
    left=2pt,
    right=2pt,
    top=2pt,
    bottom=2pt,
    arc=1pt,
    boxrule=0.3pt
]
\small
\textbf{Input:} Candidate generations $\{y_i\}$ for each multimodal sample $x$.  
\textbf{Output:} Reward-weighted gradient updates favoring perceptually grounded and coherent responses.
\end{tcolorbox}

To concretize the reward signal $\mathcal{R}(x_i, y_i)$ used in Eq. \ref{eq:rro}, we define a perceptual reward term that measures multimodal alignment between the model’s prediction and ground-truth sensory evidence:
\begin{equation}\footnotesize
r_i = \mathcal{R}(x_i, y_i) = f_{\text{perceptual}}(\text{Sync, Face, Speech, Diarzation}),
\end{equation}
where $f_{\text{perceptual}}$ aggregates consistency scores from four perceptual agents. This reflective reward ensures that feedback remains grounded in perceptual correctness rather than text-only similarity, stabilizing the RRO loop shown in Fig.~\ref{fig:raft}.

To align the model’s self-reflection with multimodal correctness, 
we introduce RRO, a lightweight generative reward. 
Instead of a separate critic, we compute sequence-level rewards 
$r_i = \mathcal{R}(x, y_i)$ from the model’s own reflective feedback 
and teacher-guided scores (e.g., grounding accuracy, speaker consistency, 
and textual coherence). For $K$ sampled candidates $\{y_i\}$, 
we consider reward weighted regression update~\cite{peters2007reinforcement,black2023training}:
%
\begin{equation}
\begin{aligned}
\nabla_\theta J_{\text{RRO}}
&= \sum_{i=1}^{K} 
    w_i \, \nabla_\theta \log \pi_\theta (y_i|x), \\
w_i 
&= 
\frac{\exp\!\big(\beta (r_i-\bar r)\big)}
     {\sum_{j} \exp\!\big(\beta (r_j-\bar r)\big)},
\end{aligned}
\label{eq:rro}
\end{equation}
%
\noindent where $\bar r$ is the average reward. Compared with the popular GRPO~\cite{shao2024deepseekmath}, which employs a linear weighting update version, we found the softmax to be more stable to train empirically.

\begin{figure*}[h]
\centering
\includegraphics[width=1\textwidth]{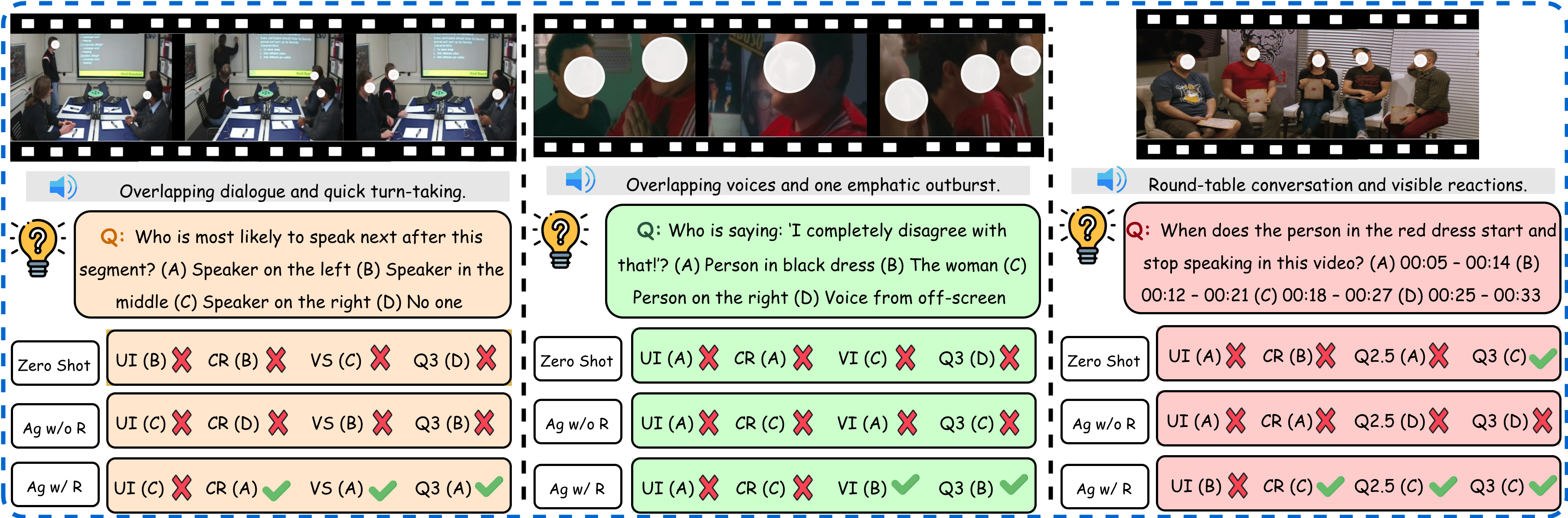}
\caption{\textbf{Qualitative results.} Comparison on multi-speaker reasoning tasks: Next-Speaker Prediction (left), Speaker Association (middle), and Temporal Grounding (right).  UI: \textit{Unified-IO2}, CR: \textit{CREMA}, VS: \textit{VideoSALMONN}, VI: \textit{VITA}, Q2.5: \textit{Qwen2.5-Omni}, and  Q3: \textit{Qwen3-Omni} under Zero-Shot, Agentic w/o \oursolution, and Agentic w/ \oursolution modes.}
\label{fig:qual}
\end{figure*}

\noindent To prevent drift across modalities during reasoning, we incorporate a 
multimodal temporal-coherence loss:
\begin{equation}\footnotesize
\mathcal{L}_{\text{temp}}
= 
\sum_{t} \left(
\|f_a(t) - f_v(t)\|_2^2 
+ 
\gamma \|f_t(t) - f_r(t)\|_2^2
\right),
\end{equation}
where $f_a, f_v, f_t, f_r$ are audio, visual, textual, and reflective embeddings 
at timestep $t$. This regularizer enforces synchronous grounding across streams.
The final \oursolution objective integrates alignment, temporal coherence, 
and reflective reward:
\begin{equation}
\mathcal{L}_{\texttt{RAFT}}
=
\mathcal{L}_{\text{align}}
+
\alpha \mathcal{L}_{\text{temp}}
-
\beta J_{\text{RRO}}.
\end{equation}

\noindent\textbf{Selective Reasoning Adaptation (SRA).}
Due to limited data and to ensure parameter efficiency, we employs SRA a targeted adapter scheme 
applied only to multimodal reasoning layers. 
for multimodal reasoning. 
Unlike generic LoRA tuning, we restrict updates 
to cross-modal inference paths, improving interpretability and convergence 
while maintaining low computational cost.

\section{Experiments and Results}
\noindent\textbf{Baselines.} Our evaluation comprehensively assesses various MLLMs, including closed-source and open-source models. We evaluate the following closed source models \textit{GPT-4o} \cite{hurst2024gpt}, \textit{GPT-4o-mini} \cite{hurst2024gpt}, \textit{REKA} \cite{reka} and the open source models: \textit{Unified-IO2} \cite{unifiedio2}, \textit{CREMA} \cite{crema}, \textit{VideoSALMONN} \cite{videosalmonn}, \textit{VITA} \cite{vita}, \textit{Qwen2.5-Omni} \cite{xu2025qwen25omnitechnicalreport}, \textit{Qwen3-Omni} \cite{qwen3omni}. We include human evaluation results to establish an upper bound for model performance, providing a subjective reference for the achievable ceiling for each downstream task. 


\vspace{1mm}
\customsubsection{Metrics}
For text generation quality, we report \textit{BLEU@4}~\cite{papineni2002bleu}, \textit{METEOR}~\cite{meteor}, and \textit{CIDEr}~\cite{cider}. For discrete classification tasks, we use \textit{Top-1 Accuracy}. Below, we describe the additional metrics used to capture temporal and reasoning-specific performance.

\noindent\textbf{GPT-as-a-Judge Evaluation.}
We adopt the GPT-as-a-Judge evaluation~\cite{xu2025qwen25omnitechnicalreport} to assess semantic correctness, factual grounding, and reasoning quality beyond lexical overlap. An LLM is given the task, ground truth, and model response, and outputs a normalized score $s \in [0,10]$ based on task-specific rubrics, offering a scalable and human-aligned measure of multimodal understanding.

\noindent\textbf{Temporal Intersection over Union (tIoU).}
For temporal grounding and localization tasks, we use tIoU to quantify alignment between predicted and ground-truth time intervals $(\hat{t}_s, \hat{t}_e)$ and $(t_s^*, t_e^*)$:
\begin{equation}
    \text{tIoU} = 
    \frac{\min(\hat{t}_e, t_e^*) - \max(\hat{t}_s, t_s^*)}
         {\max(\hat{t}_e, t_e^*) - \min(\hat{t}_s, t_s^*)}.
\end{equation}
Higher tIoU values indicate better temporal precision and consistency in identifying speaker activity boundaries.

\noindent\textbf{Off-by-One Accuracy.}
We report Off-by-One Accuracy to capture tolerance to minor temporal deviations in sequential labeling tasks. A prediction is considered correct if it falls within one step of the ground truth:
\begin{equation}
    \text{Acc}_{\text{OBO}} = 
    \frac{1}{N} \sum_{i=1}^{N} 
    \mathbbm{1}\big[|\hat{y}_i - y_i^*| \leq 1\big].
\end{equation}
This metric accounts for small timing offsets that do not affect the temporal correctness of the prediction.

\noindent\textbf{Human-Judged Coherence.}
This is a subjective metric assessing how coherently a model links events or dialogues across scenes. Human evaluators rate the logical, temporal, and contextual consistency of responses on a 0–10 scale, with higher scores indicating stronger narrative flow and cross-scene understanding.

\begin{figure}[t!]
    \centering
    \includegraphics[width=0.7\linewidth]{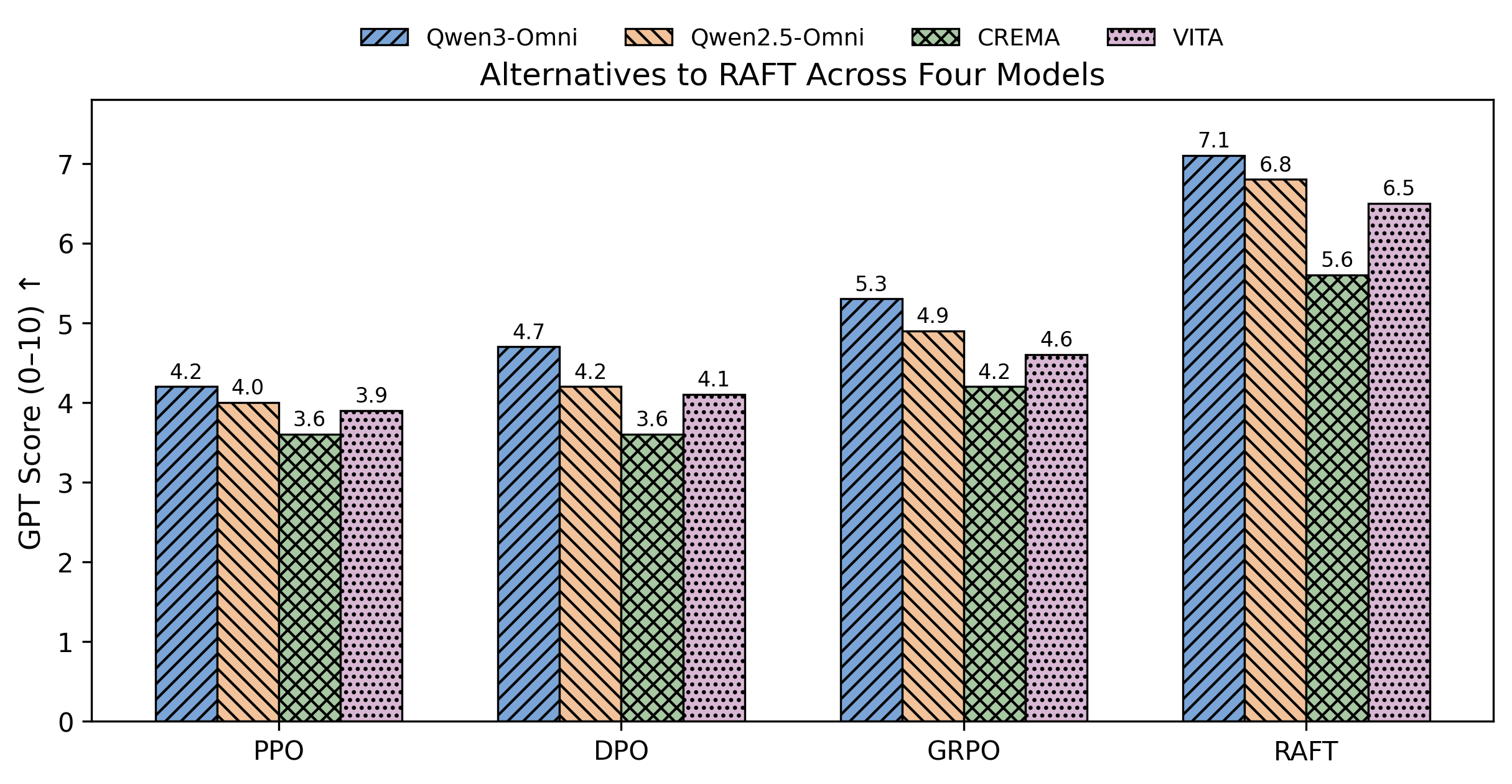}
    \caption{\textbf{Comparison of optimization methods.} Agentic performance across models on STG task.}
    \label{fig:raft-rlhf}
\end{figure}

\begin{figure}[t!]
    \centering
    \includegraphics[width=0.7\linewidth]{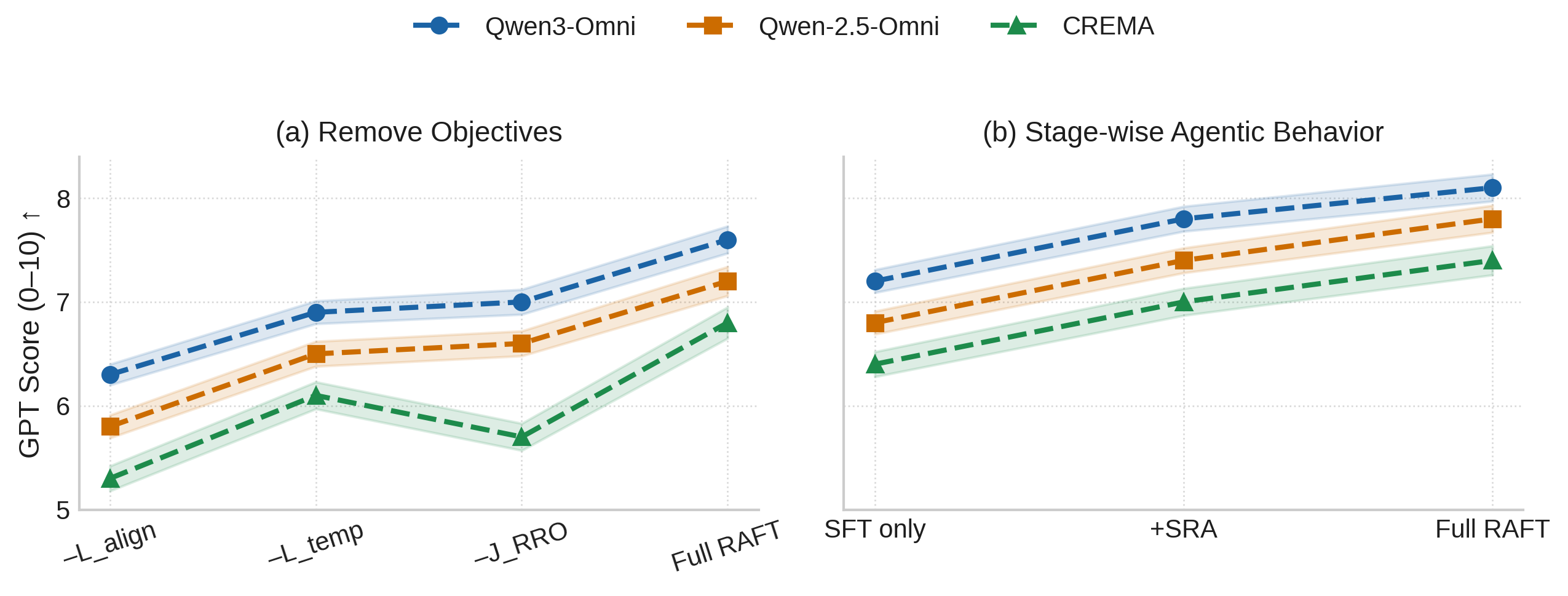}
    \caption{\textbf{Ablation analysis of \oursolution on AVDS.} (a) Removing core objectives degrades performance across models (b) Highlights stage-wise gains.
    }
    \label{fig:raft-ablation}
\end{figure}

\vspace{2mm}
\customsubsection{Findings}
\label{sec:findings}

\noindent\textcolor{blue!60!black}{\textbf{Performance declines with autonomy.}}  
Model accuracy and reasoning coherence consistently drop as autonomy increases from ZS to agentic evaluation (Tab. \ref{tab:avds}--\ref{tab:stg-csnl}). Most MLLMs struggle to maintain speaker consistency and contextual grounding once explicit cues are removed, revealing a dependence on external structure rather than internal temporal modeling.

\noindent\textcolor{orange!75!black}{\textbf{Over-reliance on guided prompts.}}  
While models such as \textit{GPT-4o} and \textit{Qwen3-Omni} perform competitively in guided modes, their reasoning quality, turn attribution, and temporal flow degrade sharply without instructions. This indicates that prompt scaffolding currently substitutes for robust multimodal representations.

\noindent\textcolor{teal!60!black}{\textbf{Task sensitivity to autonomy.}}  
Among all tasks, \textit{audio-visual summarization}, \textit{speaker temporal grounding} and \textit{cross-modal narrative linking} exhibit the steepest performance decline from guided to autonomous modes, reflecting the compounded difficulty of multimodal fusion and long-range speaker tracking. In contrast, \textit{next-speaker prediction} and \textit{speaker association} remain more stable, suggesting that short-term conversational cues are easier for present MLLMs to capture.

\noindent\textcolor{green!50!black}{\textbf{Effect of \oursolution.}}  
Integrating \oursolution\ consistently boosts performance across models and modes, yielding on average $+6.7$ BLEU, $+1.1$ METEOR, and $+6.8$ CIDEr, with GPT-based human-alignment gains up to $+1.5$. The improvements are most pronounced in autonomous settings, where structured reasoning and self-reflection reduce temporal drift and speaker confusion. Qualitatively, \oursolution-enhanced models display clearer stepwise reasoning, explicit planning (e.g., identifying speaker order before summarization), and stronger factual consistency.

\noindent\textbf{Closed-source vs. Open-source Models.}
\label{sec:closed-open}
Closed-source MLLMs consistently maintain an edge over open-source counterparts in subjective and objective metrics under zero-shot and guided conditions.  
This advantage likely stems from larger-scale audio-visual pretraining and richer cross-modal alignment.    
After \oursolution finetuning high-quality open-source models such as \textit{Qwen3-Omni} demonstrate adaptability, suggesting that open-source models possess stronger generalization when allowed to self-direct reasoning.


After \oursolution fine-tuning, multiple open-source models close and often surpass the performance gap with closed-source ones on most metrics.
This demonstrates \oursolution as a \emph{model-agnostic enhancement recipe} that unlocks agentic competence even in smaller, lightweight models.
Its adaptability makes it practical and data-efficient, achieving scalable improvement without proprietary pipelines.
Overall, structured agentic training enables open-source models to rival or exceed closed-source systems in multi-speaker reasoning, showing that \textit{agentic understanding is learnable, not inherent}.


\customsubsection{Ablation Studies}
\label{sec:ablations}


\vspace{2mm}

\noindent\textbf{Effect of Optimization Strategy}
Fig.~\ref{fig:raft-rlhf} compares PPO \cite{ppo}, DPO \cite{dpo}, GRPO \cite{grpo}, with \oursolution on the Speaker Temporal Grounding task. \oursolution achieves the highest agentic scores, showing stronger alignment of temporal and multimodal cues. While GRPO offers moderate gains, PPO and DPO lag due to weaker multimodal feedback. Notably, larger models (e.g., \textit{Qwen3-Omni}) benefit most, though smaller ones like \textit{CREMA} also improve highlighting \oursolution’s scalable and robust optimization.

\noindent\textbf{Insights from \oursolution Ablations.}
Fig.~\ref{fig:raft-ablation} shows that \oursolution’s gains arise from the complementary roles of its components. Removing alignment, temporal grounding, or reflective optimization degrades performance, with the reflective term contributing most to resolving multi-speaker ambiguities. Stage-wise results confirm that improvements emerge after adding SRA and reflection, highlighting \textit{progressive specialization and self-correction} over mere supervision.

\vspace{1mm}

\customsubsection{Qualitative Results}

Fig. \ref{fig:qual} compares models across three multi-speaker tasks. Baselines like \textit{Unified-IO2}, \textit{CREMA}, and \textit{VideoSALMONN} often fail under overlapping speech, while the agentic \textit{Qwen2.5-Omni} and \textit{Qwen3-Omni} models with \oursolution yield more coherent and temporally grounded predictions. The reflective phase helps refine speaker and timing consistency in complex dialogues.

\section{Conclusion and Future Work}
\label{sec:conclusion}
We introduced \ourbenchmark, a unified benchmark for evaluating and improving agentic multi-speaker understanding in MLLMs. It benchmarks models across zero-shot, guided, and autonomous settings, revealing their drawbacks. To address that, we proposed \oursolution, a model-agnostic approach combining structured Plan–Act–Reflect supervision with GRPO-based reward alignment. \oursolution improves reasoning consistency and speaker tracking with minimal labelled data and parameter updates. Together, they bridge evaluation and improvement, laying the foundation for more robust and socially aware multimodal agents.

Future work can aim to generalize the reward-aligned optimization principles to broader multimodal reasoning tasks. By integrating temporal alignment with reflective feedback across vision–language and embodied environments, such extensions could enable a unified framework for improved grounding and cross-modal interaction.


\newpage
\appendix
\appendixpage

\startcontents[sections]
\printcontents[sections]{l}{1}{\setcounter{tocdepth}{2}}
\newpage

\section{Supplementary Video}
\label{supp video}
In the supplementary video, we provide audio-visual examples for each task and compare the performance of different models across \textit{zero-shot}, \textit{guided} and \textit{agentic} modes of evaluation. The video also explains how the perception tools are being leveraged at different instances to improve the overall model performance.

\section{Additional Details About \ourbenchmark}
\label{data construction}

\begin{figure}[h!]
    \centering
    \includegraphics[width=0.5\linewidth]{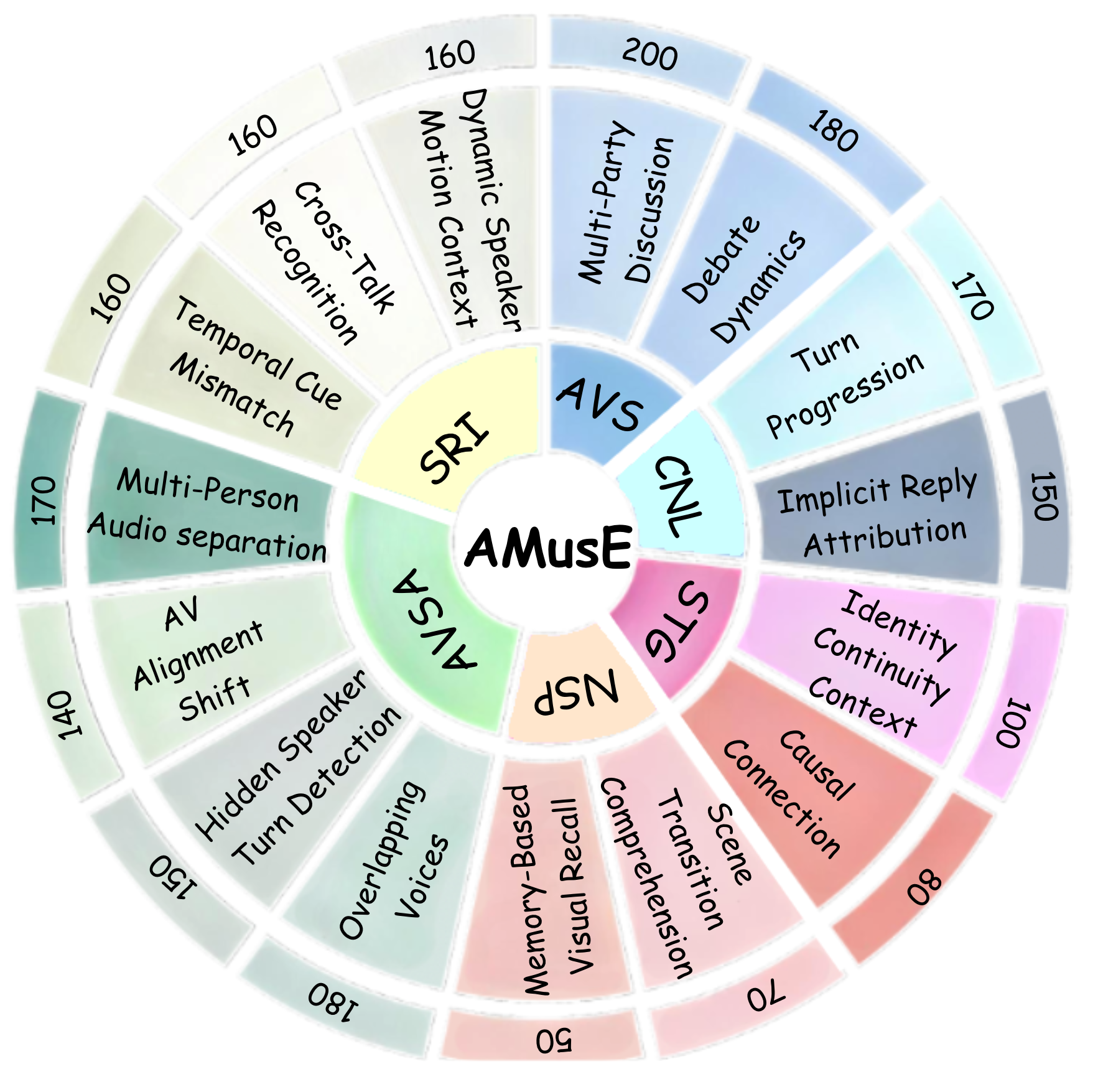}
    \caption{\textbf{\ourbenchmark taxonomy.}  Hierarchical taxonomy of 15 multi-speaker reasoning scenarios across 6 tasks highlighting the diversity and complexity of \ourbenchmark conversations.}
    \label{fig:amuse_pie}
\end{figure}

\begin{figure*}[h]
\centering
\includegraphics[width=1\textwidth]{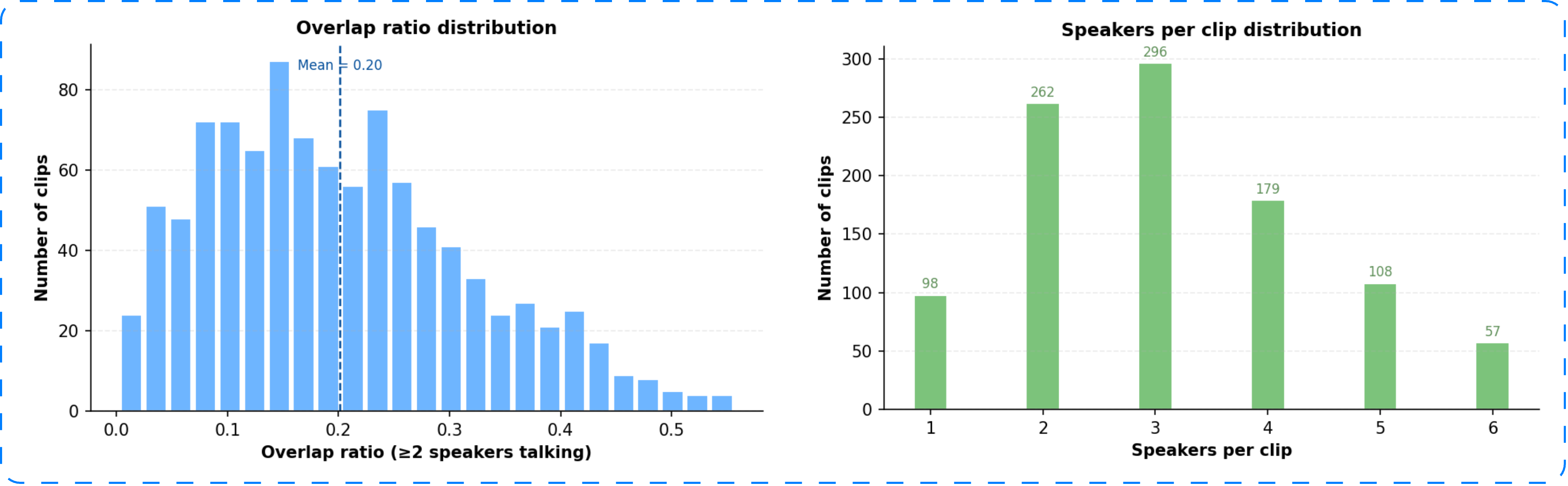}
\caption{\textbf{Dataset statistics.} Distribution of multi-speaker overlap ratios (left) and number of visible speakers per clip (right) in \ourbenchmark. The benchmark contains substantial speaker overlap and diverse group sizes, reflecting realistic multi-party conversational settings.}
\label{fig:amuse_stats-supp}
\end{figure*}

\subsection{Dataset Construction}
\ourbenchmark is constructed through an automated pipeline by leveraging the metadata from the underlying datasets. The samples are derived from the ground-truth annotations available in the source datasets: AVA Active Speaker \cite{gu2018ava}, VoxCeleb2 \cite{chung2018voxceleb2}, FriendsMMC \cite{wang2025friends}, AMI Meetings \cite{kraaij2005ami}, and curated YouTube videos with manually verified metadata. Our pipeline defines explicit rules for (i) extracting valid temporal segments, (ii) mapping them to speakers or events, (iii) stitching or pairing segments across clips when required by the task, and (iv) generating template-based queries grounded in annotated timestamps, transcripts, and speaker identities. Below we describe the construction procedures for each task.

\subsubsection{Speaker Temporal Grounding}
For each video, we extract speech-active regions using the dataset-provided temporal annotations (e.g., AVA active-speaker labels, AMI speech segments etc). A valid grounding sample is formed by selecting a continuous utterance from a target speaker and preserving its exact start and end timestamps. The transcript for this interval is obtained by slicing the dataset transcript according to the same boundaries. All timestamps originate directly from annotation files. Questions are generated using templates that reference these grounded intervals, ensuring the textual prompt aligns perfectly with the annotated speech region.

\subsubsection{Audio-Visual Dialogue Summarization}
Dialogue summarization samples are built by selecting multi-turn segments from the annotated transcripts. Each clip is formed by choosing a fixed-length temporal window (e.g., upto 40 seconds) and bundling all utterances occurring within that window. Speaker identities and utterance order are inherited from the transcript metadata. The transcript snippet is created by concatenating the turn-by-turn dialog text exactly as annotated, without modification or re-alignment. Summaries are then generated through structured templates that reference only the content contained in the selected segment.

\subsubsection{AV Speaker Association}
To build association samples, we first identify all speech segments from a given video along with their corresponding annotated speaker IDs. For each speech interval, we pair the audio-driven transcript slice with the video frames in which the same speaker appears, using the dataset-provided face track or region-of-interest metadata (e.g., bounding box indices in AVA or FriendsMMC). The pipeline directly maps the annotated speaker ID to the associated temporal window; no cross-modal embedding or recognition is involved. For each question, the template either asks the model to identify which speaker produced a given utterance or to decide whether a specific person in the scene said a particular line.

\subsubsection{Next Speaker Prediction}
This task is constructed using dialogue ordering rules. For any multi-speaker segment, we inspect the transcript metadata and detect the temporal boundary between two consecutive speakers. A training sample is created by selecting the preceding context (a sequence of utterances) and identifying the next annotated speaker as the prediction target. Only the transcript order and timestamps are used; no additional ASR or heuristic filtering is applied. If a segment contains overlapping or indistinguishable turns based on metadata alone, it is automatically discarded to maintain clean conversational structure.

\subsubsection{Speaker Re-identification}
Re-identification samples are constructed by pairing two segments of the same annotated speaker at different times within the same dataset. For each identity, we gather all timestamps where that speaker appears, select two non-overlapping intervals, and extract the corresponding transcript slices. Positive pairs reuse the same speaker ID; negative pairs are formed by pairing intervals belonging to distinct IDs. All identity information is taken strictly from the dataset's annotations (VoxCeleb2 IDs, FriendsMMC cast labels, AMI participant tags), with no external face verification. Only pairs with sufficient temporal separation or scene variation are retained to ensure meaningful samples.


\subsubsection{Cross-Scene Narrative Linking}

\textit{\underline{(i) Scene Selection and Grouping.}}
We construct CSNL samples by identifying pairs or triplets of scenes from sitcoms, drama series, talk shows, and long-form conversational videos where the narrative in one segment depends on an event in another. Instead of using perceptual tools, we rely entirely on transcript metadata, speaker IDs, and annotated timestamps to detect recurring entities, callbacks, or references across non-contiguous segments. Scenes are grouped when (i) they share an annotated speaker or participant, (ii) the transcript explicitly references an earlier event, or (iii) later dialogue resolves or reacts to information introduced in a prior segment.

\noindent \textit{\underline{(ii) Template-Based Prompt and Option Design.}}
For each grouped set of scenes, annotators write a question that explicitly requires linking the two temporally disjoint narrative events (e.g., ``Why does the person in the red sweater react that way at the end of the video?''). All questions are grounded in the transcript segments extracted directly from timestamped annotations. Annotators then construct four answer choices (one correct, three distractors), ensuring that the correct answer can only be obtained by integrating information across scenes rather than relying on local, scene-specific cues. Distractors are crafted to be plausible based on individual scenes but incorrect when cross-scene reasoning is applied.

\noindent \textit{\underline{(iii) Coherence and Difficulty Verification.}}
An annotator evaluates each CSNL item for narrative coherence, cross-scene consistency, and difficulty using a Likert-scale rating. Items receiving low coherence scores, containing ambiguous distractors, or failing to require explicit cross-scene reasoning are removed. The final CSNL set therefore includes only those samples where the narrative link is unambiguous, grounded entirely in the annotated transcripts, and cannot be solved without integrating information across multiple scenes.

\noindent\textbf{Rule-Based Engineering Pipeline.}
Across all tasks, the pipeline is governed by deterministic rules:
(i) segments are extracted strictly using timestamp metadata;
(ii) transcripts are produced by slicing and concatenating ground-truth tokens;
(iii) speaker IDs originate solely from dataset annotations;
(iv) multi-segment samples are stitched by pairing annotated intervals without any perceptual inference; and
(v) every query is produced by a template that references only validated events, speakers, and timestamps.

\noindent\textbf{Final Curation.}
We apply checks to ensure that the stitched intervals, paired speakers, and extracted transcripts are internally consistent (matching IDs, non-overlapping timestamps, correct temporal ordering). Ambiguous or borderline segments are pruned directly from the metadata rather than post-processed. This ensures \ourbenchmark remains a clean, annotation-driven benchmark that reflects the structure and reliability of its underlying source datasets.

\subsection{Dataset Breakdown}

\ourbenchmark integrates clips from AVA Active Speaker, VoxCeleb2, FriendsMMC, AMI Meetings, and curated YouTube videos to build six multimodal reasoning tasks. As summarized in Tab. \ref{tab:dataset-distribution}, the first five tasks contain 400 samples each, while Cross-Scene Narrative Linking provides 100 multi-segment examples. Different datasets contribute complementary strengths AVA for active-speaker cues, VoxCeleb2 for identity-focused cases, FriendsMMC and AMI for rich multi-party dialogue, and YouTube for unconstrained scenarios. The dataset exhibits diverse speaker dynamics, with overlap ratios centered around 0.20 and clips containing 1–6 visible speakers (Fig. \ref{fig:amuse_stats-supp}). The semantic wheel (Fig. \ref{fig:amuse_pie}) further shows that each task targets a distinct reasoning challenge, spanning grounding, turn-taking, association, identity persistence, and narrative linkage.

\begin{table}[t]
\centering
\small
\setlength{\tabcolsep}{6pt}
\resizebox{0.7\linewidth}{!}{
\begin{tabular}{lccccc}
\toprule
\textbf{Task} & \textbf{AVA} & \textbf{VoxCeleb2} & \textbf{FriendsMMC} & \textbf{AMI} & \textbf{YouTube} \\
\midrule
Speaker Temporal Grounding & 120 & 80 & 80 & 90 & 30 \\
AV Dialogue Summarization  & 100 & 70 & 90 & 110 & 30 \\
AV Speaker Association     & 130 & 90 & 70 & 80 & 30 \\
Next Speaker Prediction    & 110 & 80 & 100 & 80 & 30 \\
Speaker Re-identification  & 90  & 140 & 60 & 80 & 30 \\
Cross-Scene Narrative Linking & 0 & 0 & 70 & 0 & 30 \\
\midrule
\textbf{Total} & 550 & 460 & 470 & 440 & 180 \\
\bottomrule
\end{tabular}}
\caption{Number of samples collected from each dataset for all six \ourbenchmark tasks. The first five tasks contain 400 samples each, while Cross-Scene Narrative Linking contains 100 samples. Counts reflect how clips were sourced from AVA Active Speaker, VoxCeleb2, FriendsMMC, AMI Meetings, and curated YouTube videos.}
\label{tab:dataset-distribution}
\end{table}

\subsection{Quality Control}

\subsubsection{Speaker Temporal Grounding}
For each sample, we verify that the start and end timestamps fall strictly within valid speech-activity regions annotated in the source dataset. We also confirm that the speaker identity associated with the grounded span matches both the diarization labels and the face track present in the corresponding frames. Samples with boundary mismatches, off-by-one frame shifts, or incorrect speaker assignments are flagged and corrected by re-aligning timestamps using dataset-specific metadata (e.g., AVA Active Speaker \cite{gu2018ava} labels or AMI \cite{kraaij2005ami} segment boundaries).

\subsubsection{Audio-Visual Dialogue Summarization}
Each summary query is checked for semantic consistency with the transcript. We use a two-step validator: (i) template-level checks to ensure the summary references only events that occur within the clip, and (ii) LLM-based semantic alignment scoring to flag summaries that omit key events or hallucinate nonexistent content. Misaligned samples are regenerated by re-running the summarization template or corrected through constrained editing.

\subsubsection{AV Speaker Association}
We verify that the utterance-to-speaker mapping is consistent across modalities. Each audio segment is matched against the active face track using cross-modal similarity checks, which include active-speaker labels (AVA), speaker-ID metadata (VoxCeleb2 \cite{chung2018voxceleb2}), and video-based face clustering. Mismatches, such as a voice segment mapped to the wrong face, are automatically flagged using speaker-consistency rules and reprocessed by reassigning the correct identity or discarding ambiguous segments.

\subsubsection{Next Speaker Prediction}
For next-speaker queries, we inspect the dialogue ordering by validating that the predicted ``next'' speaker is temporally the next annotated speaker in the transcript and visually present in the upcoming frames. We also verify that no overlap, interruption, or missing segment disrupts the dialogue sequence. Samples with inconsistent ordering or missing participants are corrected through timeline realignment or regenerated from cleaner clips.

\subsubsection{Speaker Re-identification}
To ensure identity consistency across disjoint segments, each query pair is validated by comparing speaker embeddings from the original source dataset (e.g., VoxCeleb2 \cite{chung2018voxceleb2} identity embeddings, FriendsMMC \cite{wang2025friends} cast labels, or AMI participant IDs). We confirm that the positive pairs indeed correspond to the same canonical identity, and that negative pairs have no shared speaker ID. Any identity drift (e.g., due to noisy face tracks or brief occlusions) is rectified by reassigning IDs or replacing the clip.

\subsubsection{Cross-Scene Narrative Linking}
Given the reasoning-heavy nature of this task, we perform additional checks to ensure narrative coherence between scenes. The validator ensures that the cross-scene question references entities or events that exist in both clips according to the annotated transcripts. We also use semantic matching to ensure that the link between scenes (e.g., shared speaker, consistent topic, or causal relation) is grounded in the metadata. Incorrect or weakly-connected samples are regenerated with stricter constraints on shared entities or topics.

\noindent\textbf{Final Sanity Checks.}
Across all tasks, we perform automated mismatch detection to identify: (i) incorrect speaker labels, (ii) misaligned timestamps, (iii) invalid temporal spans, (iv) hallucinated entities in template-generated questions, and (v) multi-modal inconsistencies between audio, visual tracks, and transcripts. Detected issues are rectified either by re-extracting metadata from the raw sources, regenerating templates with stricter constraints, or manual inspection for borderline cases. This ensures that every \ourbenchmark sample is temporally accurate, speaker-consistent, and semantically grounded.


\begin{table*}[t]
\centering
\small
\renewcommand{\arraystretch}{1.2}
\setlength{\tabcolsep}{5pt}
\begin{tabular}{p{0.16\linewidth} p{0.27\linewidth} p{0.27\linewidth} p{0.24\linewidth}}
\toprule
\textbf{Tool} & \textbf{Inputs} & \textbf{Outputs} & \textbf{Example API-style call} \\
\midrule

\textbf{Whisper} &
Audio file (wav/mp3); optional language tag. &
Timestamped transcript; word segments; confidence scores. &
\scriptsize asr = whisper.transcribe("clip.wav", model="large") \\[0.3em]

\textbf{Pyannote} &
Audio waveform; diarization pipeline instance. &
List of (start, end, speaker\_id); overlap flags. &
\scriptsize dia = pipeline("speaker-diarization")("clip.wav") \\[0.3em]

\textbf{InsightFace} &
Video frames or mp4 file. &
Face detections; embeddings; track IDs; bounding boxes with timestamps. &
\scriptsize faces \eq insightface.get\_faces("clip.mp4") \\[0.3em]

\textbf{SyncNet} &
Video file and audio track. &
AV sync offset; confidence score; mapping from speech segments to face tracks. &
\scriptsize sync = syncnet.align(video="clip.mp4", audio="clip.wav") \\

\bottomrule
\end{tabular}
\caption{Inputs, outputs, and example API-style calls for the four perception tools used in \ourbenchmark: Whisper (ASR), PyAnnote (speaker diarization), InsightFace (face tracking/recognition), and SyncNet (audio–visual synchronization). }
\label{tab:tool-api-snippets}
\end{table*}

\subsection{Question Templates}
Here we add question template for each task in Tab. \ref{tab: question avds} - Tab. \ref{tab: question csnl}. Each task in \ourbenchmark is paired with a diverse set of carefully designed question templates that capture the core reasoning abilities required for that task. For example, Audio-Visual Dialogue Summarization includes prompts that ask annotators or models to restate, condense, or paraphrase a speaker’s message within a specified temporal segment. Similarly, the other five tasks Speaker Temporal Grounding, AV Speaker Association, Next Speaker Prediction, Speaker Re-identification, and Cross-Scene Narrative Linking each use their own bank of structured question variants tailored to highlight temporal localization, identity matching, conversational dynamics, or narrative inference. Across tasks, these templates ensure broad semantic coverage, reduce prompt bias, and provide consistent evaluation signals while reflecting the natural variability of real-world multimodal queries.




\section{Post-Training Intuition}
\label{post training intuition}

While prior multimodal alignment methods rely on policy optimization (e.g., GRPO) or lightweight adapter tuning (e.g., LoRA), \oursolution restructures post-training around three mathematically grounded components:

\subsection{Self-Reflective Rewarding}
    For a multimodal input 
    $x = (x^{(a)}, x^{(v)}, x^{(t)})$ 
    and model output $(\hat{y}, r)$ consisting of an answer and a reasoning trace, we define the intrinsic reward as
   \begin{equation}
\label{eq:self_reflective_reward}
\begin{aligned}
R(x, \hat{y}, r)
&= \lambda_{\text{task}}\, s_{\text{task}}(\hat{y}) \\
&\quad + \lambda_{\text{align}}\, s_{\text{align}}(x, r) \\
&\quad + \lambda_{\text{conf}}\, s_{\text{conf}}(\hat{y}) .
\end{aligned}
\end{equation}
    Here, $s_{\text{task}}$ measures task-level correctness (e.g., answer accuracy or span IoU), 
    $s_{\text{align}}$ captures cross-modal consistency between referenced speakers/timestamps and the underlying audio–visual evidence, 
    and $s_{\text{conf}}$ penalizes unsupported over-confident predictions.
    All components are derived from the model’s internal probabilities and alignment scores no external reward model is used.
    Intuitively, the model is rewarded when its predictions and explanations agree with the multimodal input and penalized when they contradict themselves.

\subsection{Selective Reasoning Adaptation}
    We decompose model parameters into 
    $\theta = (\theta_{\text{base}},\,\theta_{\text{cross}})$, 
    where $\theta_{\text{cross}}$ corresponds to explicitly interpretable cross-modal reasoning blocks.
    During training, we mask gradients such that only cross-modal parameters are updated:
    \begin{align}
        \tilde{\nabla}_{\theta_i}\mathcal{L}
        &= 
        \begin{cases}
            \nabla_{\theta_i}\mathcal{L}, & \theta_i \in \theta_{\text{cross}},\\[0.3em]
            0, & \theta_i \in \theta_{\text{base}}.
        \end{cases}
        \label{eq:selective_update}
    \end{align}
    This focuses adaptation on the specific components responsible for audio–visual–text reasoning, improving compute and data efficiency while avoiding catastrophic forgetting.

\subsection{Temporal Coherence Constraint.}
    Let $h^{(a)}_t$, $h^{(v)}_t$, and $h^{(t)}_t$ denote audio, visual, and text embeddings extracted at time index $t$. 
    We impose a temporal coherence loss:

\begin{equation}
\label{eq:temporal_coherence}
\begin{aligned}
\mathcal{L}_{\text{temp}}
&= \sum_{t} \Big(
      \|h^{(a)}_t - h^{(v)}_t\|_2^2
    + \|h^{(v)}_t - h^{(t)}_t\|_2^2
    \Big) \\
&\quad + \sum_{t} \gamma \Big\|
      \big(h^{(a)}_{t+1} - h^{(a)}_{t}\big)
      - \big(h^{(v)}_{t+1} - h^{(v)}_{t}\big)
    \Big\|_2^2 .
\end{aligned}
\end{equation}
    The first two terms encourage cross-modal agreement at each timestep, while the final term enforces consistent evolution over time. 
    Intuitively, this prevents identity jumps, speaker drift, or abrupt modality mismatches critical for multi-speaker tracking, next-speaker prediction, and narrative coherence.

Together, these components provide a principled post-training paradigm in which \oursolution optimizes a self-reflective reward, updates only interpretable cross-modal reasoning pathways, and maintains temporally coherent multimodal embeddings yielding efficient, stable, and grounded multimodal alignment.




\begin{table}[t]
\centering
\small
\renewcommand{\arraystretch}{0.8}
\setlength{\tabcolsep}{6pt}
\resizebox{0.6\columnwidth}{!}{
\begin{tabular}{l c c c}
\toprule
\textbf{STG} & \textbf{Metric} & \textbf{Value} & \textbf{Final IoU↑} \\
\midrule
Whisper & WER ↓ & 1.33 & \multirow{3}{*}{\centering 51.02} \\
Pyannote & DER ↓ & 1.23 & \\
SyncNet & Sync Error ↓ & 2.12 & \\
\bottomrule
\end{tabular}}
\caption{Tool-level metrics and final STG performance for Qwen3-Omni.}
\label{tab:stg-tool-supp}
\end{table}

\begin{table}[t]
\centering
\small
\renewcommand{\arraystretch}{1.15}
\setlength{\tabcolsep}{6pt}
\resizebox{0.6\columnwidth}{!}{
\begin{tabular}{l c c c}
\toprule
\textbf{AVS} & \textbf{Metric} & \textbf{Value} & \textbf{Final BLEU↑} \\
\midrule
Whisper & WER ↓ & 1.15 & \multirow{3}{*}{\centering 48.08} \\
Pyannote & Turn Acc ↑ & 89.29 & \\
InsightFace & ID Consistency ↑ & 90.33 & \\
\bottomrule
\end{tabular}}
\caption{Tool-level metrics and final AVS performance for Qwen3-Omni.}
\label{tab:avs-tool-supp}
\end{table}

\begin{table}[t]
\centering
\small
\renewcommand{\arraystretch}{1.15}
\setlength{\tabcolsep}{6pt}
\resizebox{0.6\columnwidth}{!}{
\begin{tabular}{l c c c}
\toprule
\textbf{AVSA} & \textbf{Metric} & \textbf{Value} & \textbf{Final Acc↑} \\
\midrule
Whisper & Utterance Match ↑ & 92.93 & \multirow{4}{*}{\centering 46.98} \\
Pyannote & Speaker-ID Match ↑ & 94.22 & \\
InsightFace & Face-ID Match ↑ & 90.37 & \\
SyncNet & AV Sync ↑ & 94.10 & \\
\bottomrule
\end{tabular}}
\caption{Tool-level metrics and final AV Speaker Association performance for Qwen3-Omni.}
\label{tab:avsa-tool-supp}
\end{table}

\begin{table}[t]
\centering
\small
\renewcommand{\arraystretch}{1.15}
\setlength{\tabcolsep}{6pt}
\resizebox{0.6\columnwidth}{!}{
\begin{tabular}{l c c c}
\toprule
\textbf{NSP} & \textbf{Metric} & \textbf{Value} & \textbf{Final Acc↑} \\
\midrule
Whisper & Context Coverage ↑ & 91.03 & \multirow{3}{*}{\centering 45.02} \\
Pyannote & Turn Ordering ↑ & 92.89 & \\
InsightFace & Visual ID Consistency ↑ & 93.20 & \\
\bottomrule
\end{tabular}}
\caption{Tool-level metrics and final Next Speaker Prediction performance for Qwen3-Omni.}
\label{tab:nsp-tool-supp}
\end{table}

\begin{table}[t]
\centering
\small
\renewcommand{\arraystretch}{1.15}
\setlength{\tabcolsep}{6pt}
\resizebox{0.6\columnwidth}{!}{
\begin{tabular}{l c c c}
\toprule
\textbf{SRI} & \textbf{Metric} & \textbf{Value} & \textbf{Final Acc↑} \\
\midrule
Whisper & Utterance Match ↑ & 93.27 & \multirow{3}{*}{\centering 58.65} \\
Pyannote & Identity Stability ↑ & 91.04 & \\
InsightFace & Embedding Match ↑ & 94.88 & \\
\bottomrule
\end{tabular}}
\caption{Tool-level metrics and final SRI performance for Qwen3-Omni.}
\label{tab:sri-tool-supp}
\end{table}

\begin{table}[t]
\centering
\small
\renewcommand{\arraystretch}{1.15}
\setlength{\tabcolsep}{6pt}
\resizebox{0.6\columnwidth}{!}{
\begin{tabular}{l c c c}
\toprule
\textbf{CSNL} & \textbf{Metric} & \textbf{Value} & \textbf{Final Acc↑} \\
\midrule
Whisper & Transcript Match ↑ & 89.37 & \multirow{3}{*}{\centering 49.76} \\
Pyannote & Speaker Attribution ↑ & 90.04 & \\
InsightFace & Identity Continuity ↑ & 94.11 & \\
\bottomrule
\end{tabular}}
\caption{Tool-level metrics and final CSNL performance for Qwen3-Omni.}
\label{tab:csnl-tool-supp}
\end{table}

\section{Additional Experiments}
\label{more experiments}

\subsection{\oursolution Ablations and Robustness}

\subsubsection{Effect of Temporal Regularization}

To isolate the contribution of the temporal regularizer
$\mathcal{L}_{\text{temp}}$, we ablate it from the \oursolution objective and
retrain the models. Tab. ~\ref{tab:temp-ablation} shows results on
Speaker Temporal Grounding (STG) task; similar trends hold for other
tasks.

\begin{table}[t]
\centering
\small

\begin{tabular}{lccc}
\toprule
\textbf{Model} & \textbf{Full \oursolution} & \textbf{w/o $\mathcal{L}_{\text{temp}}$} & \textbf{$\Delta$} \\
\midrule
Qwen3-Omni       & 56.3 & 51.5 & $-4.8$ \\
Qwen2.5-Omni     & 54.6 & 48.2 & $-6.4$ \\
CREMA     & 41.0 & 37.5 & $-3.5$ \\
\bottomrule
\end{tabular}
\caption{Ablation of the temporal regularizer $\mathcal{L}_{\text{temp}}$ on STG.
We report Temporal IoU (higher is better).}
\label{tab:temp-ablation}
\end{table}

\subsubsection{Softmax Temperature in RRO}

\begin{figure}[t]
    \centering
    \includegraphics[width=0.6\linewidth]{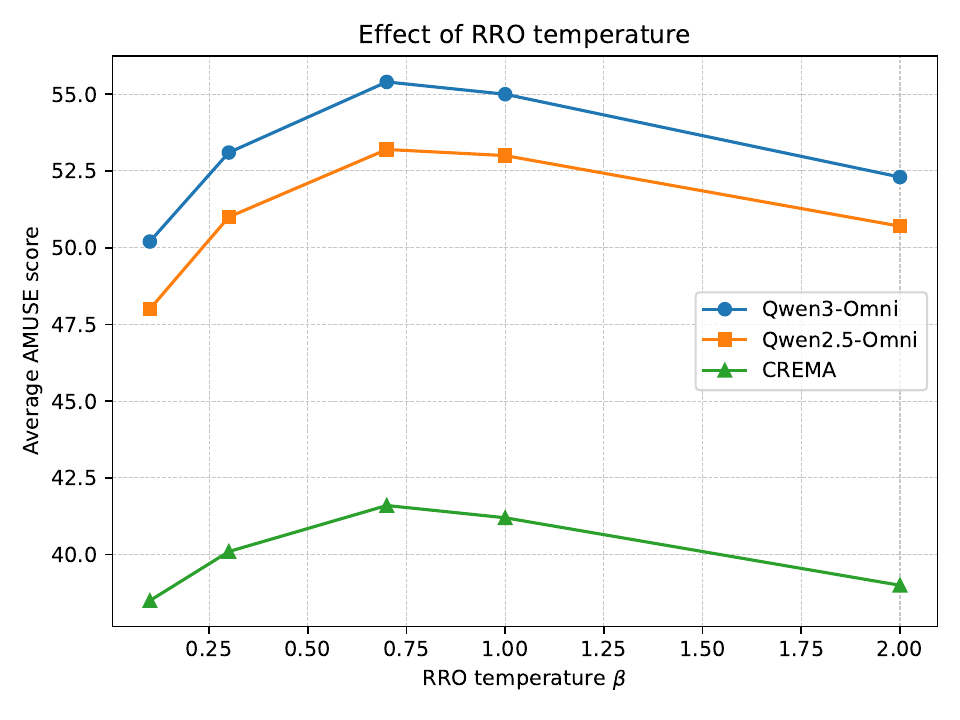}
    \caption{Effect of the RRO temperature $\beta$ on average \ourbenchmark
    performance. Extremely low or high values hurt stability, whereas
    a moderate range $\beta \in [0.3, 1.0]$ yields robust gains.}
    \label{fig:beta-sensitivity}
\end{figure}

Fig. \ref{fig:beta-sensitivity} illustrates the effect of the RRO temperature $\beta$ on average \ourbenchmark performance. Extremely low or high values of $\beta$ reduce stability by either under-emphasizing or overly sharpening the reward distribution. In contrast, a moderate range ($\beta \in [0.3,1.0]$) consistently yields higher scores across Qwen3-Omni, Qwen2.5-Omni, and CREMA, indicating that RRO benefits from controlled softmax weighting that strengthens perceptual correctness without amplifying noise.

\subsubsection{Parameter Efficiency of SRA}

\begin{figure}[t]
    \centering
    \includegraphics[width=0.6\linewidth]{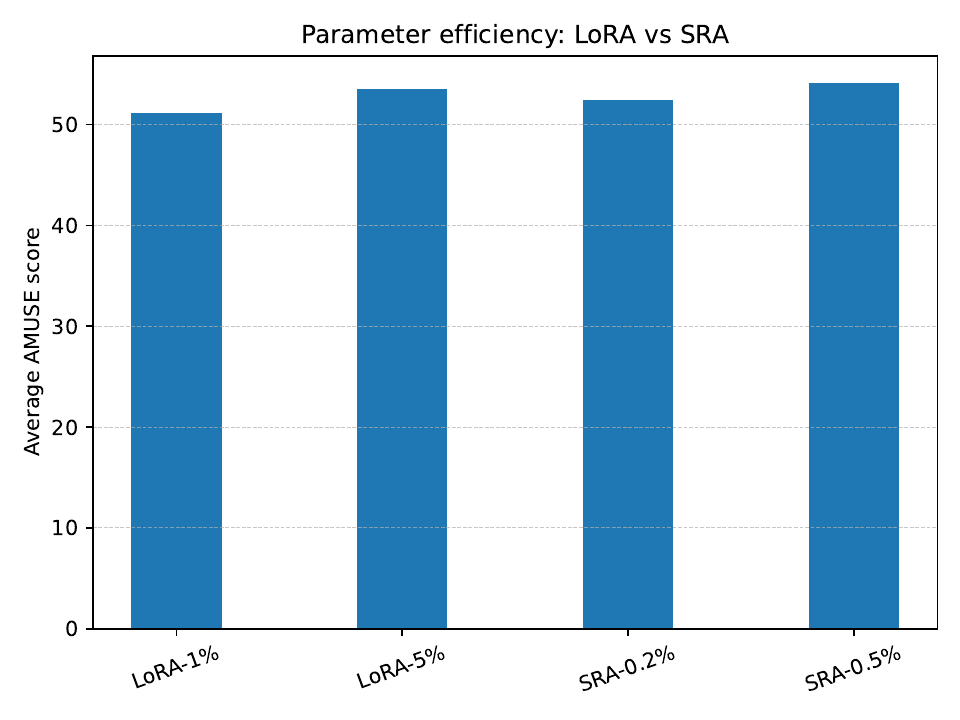}
    \caption{Average \ourbenchmark score vs.\ fraction of trainable parameters
    for LoRA and SRA on Qwen3-Omni. \oursolution with SRA achieves higher performance at
    significantly lower parameter budgets.}
    \label{fig:param-efficiency-supp}
\end{figure}

We compare \oursolution with SRA to generic low-rank adaptation (LoRA) under
different budgets of trainable parameters in Fig. \ref{fig:param-efficiency-supp}. We measure the percentage of
fine-tuned parameters relative to the backbone model. 

Tab. \ref{tab:sra-efficiency} compares LoRA and SRA under different trainable–parameter budgets. 
SRA achieves comparable or higher average scores while using an order of magnitude fewer parameters. 
Notably, SRA-0.5\% attains the best performance (54.1) despite training far fewer parameters than LoRA-5\%.

\begin{table}[t]
\centering
\small
\begin{tabular}{lccc}
\toprule
\textbf{Method} & \textbf{Trainable} \% & \textbf{Avg.\ Score} & \textbf{Rel.\ $\Delta$} \\
\midrule
LoRA-1\%   & 1.0 & 51.2 & -- \\
LoRA-5\%   & 5.0 & 53.5 & +4.5 \\
SRA-0.2\%  & 0.2 & 52.4 & +2.3 \\
SRA-0.5\%  & 0.5 & \textbf{54.1} & +5.7 \\
\bottomrule
\end{tabular}
\caption{Performance vs.\ trainable parameter budget.
SRA matches or exceeds LoRA with an order of magnitude fewer
parameters.}
\label{tab:sra-efficiency}
\end{table}

\subsection{Dataset-Centric Analyses}

\subsubsection{Speaker Overlap Difficulty}
\begin{figure}[t]
    \centering
    \includegraphics[width=0.6\linewidth]{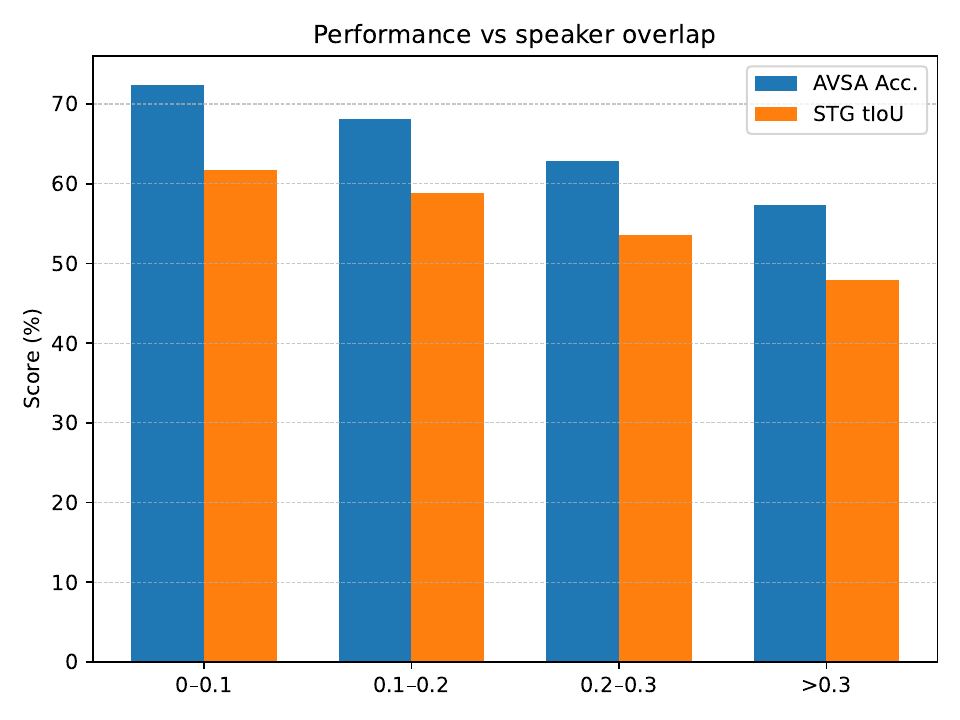}
    \caption{Performance as a function of speaker overlap ratio.
    Multi-speaker overlap substantially challenges even \oursolution-trained
    models.}
    \label{fig:overlap-difficulty-supp}
\end{figure}

Fig. ~\ref{fig:overlap-difficulty-supp} shows how speaker overlap affects performance on AVSA and STG. As the proportion of time with two or more concurrent speakers increases, both tasks exhibit a clear performance drop, highlighting the difficulty of reasoning under dense multi-speaker interactions. Even with \oursolution training, which improves grounding and temporal consistency, high-overlap scenarios remain challenging due to rapid turn-taking, overlapping utterances, and visual occlusions.



\subsubsection{Number of Visible Speakers}
\begin{figure}[t]
    \centering
    \includegraphics[width=0.6\linewidth]{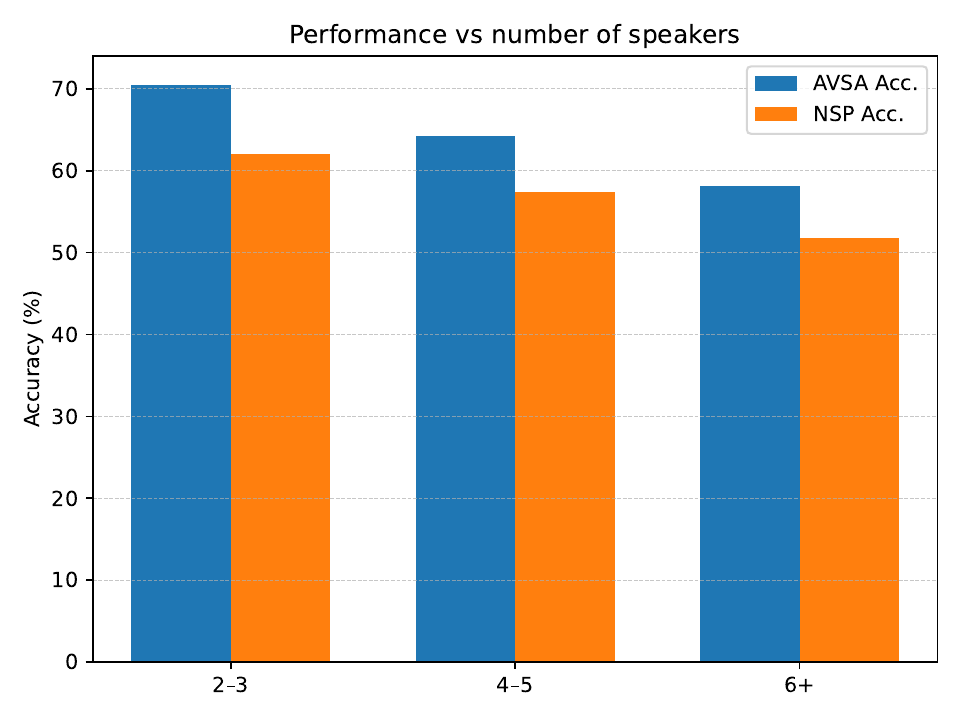}
    \caption{Accuracy vs.\ number of visible speakers on AVSA and NSP.
    Performance drops as the scene becomes more crowded.}
    \label{fig:speakers-accuracy}
\end{figure}

Fig. ~\ref{fig:speakers-accuracy} reports accuracy as a function of the number of visible speakers for AVSA and NSP. Both tasks show a consistent decline as scenes become more crowded, indicating the increased difficulty of tracking conversational roles and anticipating turn-taking when multiple participants are simultaneously visible. Even with \oursolution, higher speaker density introduces more visual competition, overlapping cues, and ambiguous interaction patterns, which collectively reduce model accuracy.


\subsubsection{Clip Duration vs.\ Accuracy}
Tab.\ref{tab:duration} reports \ourbenchmark performance as a function of clip duration. We observe a gradual decrease in accuracy as clips become longer, with the highest scores on short segments (0--20s) and the lowest on clips exceeding 40s. This trend reflects the increased reasoning difficulty in longer interactions, where models must track more speaker turns, maintain cross-modal coherence, and handle greater temporal dependencies.

\begin{table}[t]
\centering
\small
\begin{tabular}{lccc}
\toprule
\textbf{Duration} & \textbf{0--20s} & \textbf{20--40s} & \textbf{$>$40s} \\
\midrule
Avg.\ Score & 56.8 & 54.2 & 49.7 \\
\bottomrule
\end{tabular}
\caption{Effect of clip duration on average \ourbenchmark performance.}
\label{tab:duration}
\end{table}

\subsection{Agentic Tool-Use and Cue Ablations}

\subsubsection{Tool Invocation Behavior}

In agentic mode, the model decides when to call ASR (Whisper),
speaker diarization (Pyannote), and face recognition (InsightFace). We report the performance of the tool call in Tab. \ref{tab:stg-tool-supp} - Tab. \ref{tab:csnl-tool-supp}
We report the fraction of examples in which the tool decisions match
our oracle configuration in Tab. \ref{tab:tool-usage}.

\begin{table*}[t]
\centering
\small

\begin{tabular}{lccc}
\toprule
\textbf{Task} & \textbf{ASR Decision} & \textbf{Diarization Decision} & \textbf{Face-Track Decision} \\
\midrule
AVDS & 92.1 & 88.4 & 75.6 \\
AVSA & 89.3 & 94.7 & 91.2 \\
NSP  & 81.5 & 78.6 & 69.4 \\
STG  & 87.9 & 96.1 & 93.2 \\
\bottomrule
\end{tabular}
\caption{Tool selection correctness in \oursolution agentic mode.}
\label{tab:tool-usage}
\end{table*}

\subsubsection{Modality and Cue Ablations}
\begin{figure}[t]
    \centering
    \includegraphics[width=0.6\linewidth]{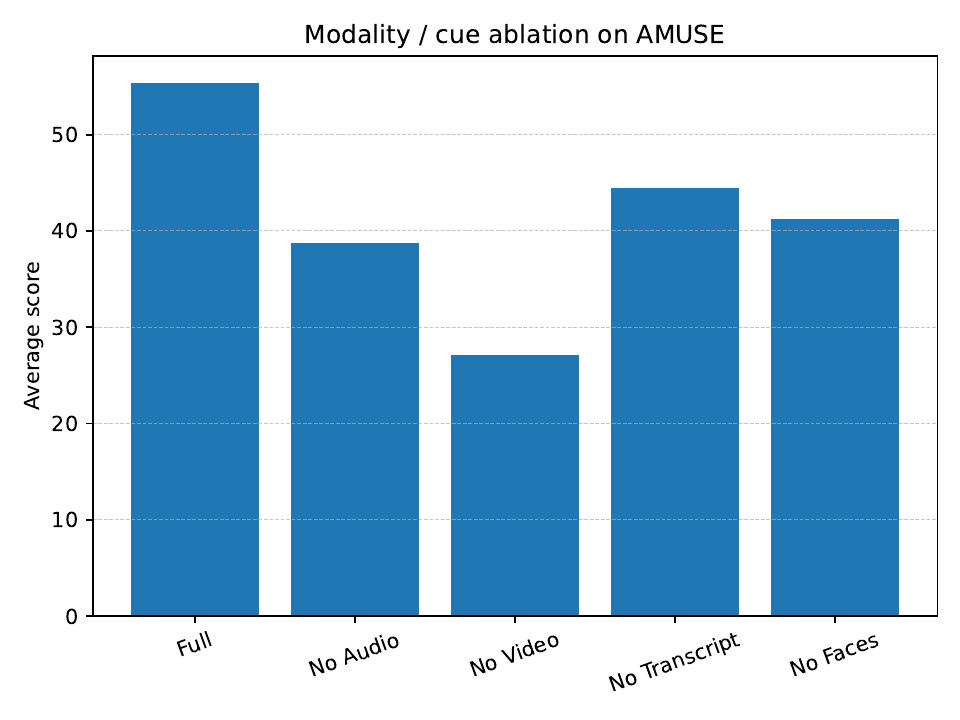}
    \caption{Effect of removing modalities and cues on average \ourbenchmark
    performance. Multi-speaker reasoning is strongly multi-modal.}
    \label{fig:cue-ablation}
\end{figure}

Fig. \ref{fig:cue-ablation} and Tab. \ref{tab:cue-ablation} present the effect of removing individual modalities and cues on average \ourbenchmark performance. Removing audio or video causes the largest degradation, confirming that multi-speaker reasoning is fundamentally audio-visual and cannot be solved from transcripts. Eliminating transcripts or face crops also reduces accuracy, though to a lesser extent, indicating that all modalities contribute complementary cues. Together, these results highlight the strongly multi-modal nature of \ourbenchmark and the importance of maintaining synchronized audio, visual, and textual information for robust reasoning.

\begin{table}[t]
\centering
\small
\begin{tabular}{lcc}
\toprule
\textbf{Setting} & \textbf{Avg.\ Score} & \textbf{$\Delta$ vs.\ Full} \\
\midrule
Full (A+V+T+F)      & 55.4 & -- \\
No Audio            & 42.7 & $-12.7$ \\
No Video            & 39.1 & $-16.3$ \\
No Transcript       & 48.5 & $-6.9$  \\
No Face Crops       & 50.2 & $-5.2$  \\
\bottomrule
\end{tabular}
\caption{Cue ablation on \ourbenchmark (average score across tasks).}
\label{tab:cue-ablation}
\end{table}

\subsection{Comparison to Other Alignment Objectives}
Tab. \ref{tab:rlhf-comparison-supp} compares \oursolution with standard RLHF approaches (PPO, DPO, and GRPO) across all six \ourbenchmark tasks on Qwen3-Omni. \oursolution consistently achieves the highest performance in every task, with especially strong gains in AVDS, NSP, SRID, and CSNL. These improvements highlight \oursolution’s ability to provide more stable reward weighting, better multimodal grounding, and more coherent step-by-step reasoning than existing preference-optimization methods.

\begin{table*}[t]
\centering
\small
\begin{tabular}{lcccccc}
\toprule
\textbf{Method} &
\textbf{AVDS (B@4)} $\uparrow$ & \textbf{AVSA (Acc\%)} $\uparrow$ & \textbf{NSP (Acc\%)} $\uparrow$ & \textbf{SRID (Acc\%)} $\uparrow$ & \textbf{STG (Acc\%)} $\uparrow$ & \textbf{CSNL (Acc\%)} $\uparrow$ \\
\midrule
PPO   & 34.82 & 61.13 & 53.28 & 58.02 & 49.44 & 39.26 \\
DPO   & 35.47 & 60.73 & 54.57 & 59.10 & 50.11 & 40.05 \\
GRPO  & 36.77 & 62.23 & 55.49 & 60.34 & 51.84 & 41.35 \\
\cellcolor{cyan!15}\textbf{\oursolution (ours)} & \cellcolor{cyan!15}\textbf{54.54} & \cellcolor{cyan!15}\textbf{54.22} & \cellcolor{cyan!15}\textbf{56.73}
            & \cellcolor{cyan!15}\textbf{62.53} & \cellcolor{cyan!15}\textbf{56.33} & \cellcolor{cyan!15}\textbf{57.26} \\
\bottomrule
\end{tabular}
\caption{Comparison of \oursolution with PPO, DPO, and GRPO across \ourbenchmark tasks on Qwen3-Omni.
We report task-specific metrics. B@4: BLEU score. }
\label{tab:rlhf-comparison-supp}
\end{table*}



\section{Additional Qualitative Results}
\label{more qual results}

\begin{figure*}[h]
\centering
\includegraphics[width=1\textwidth]{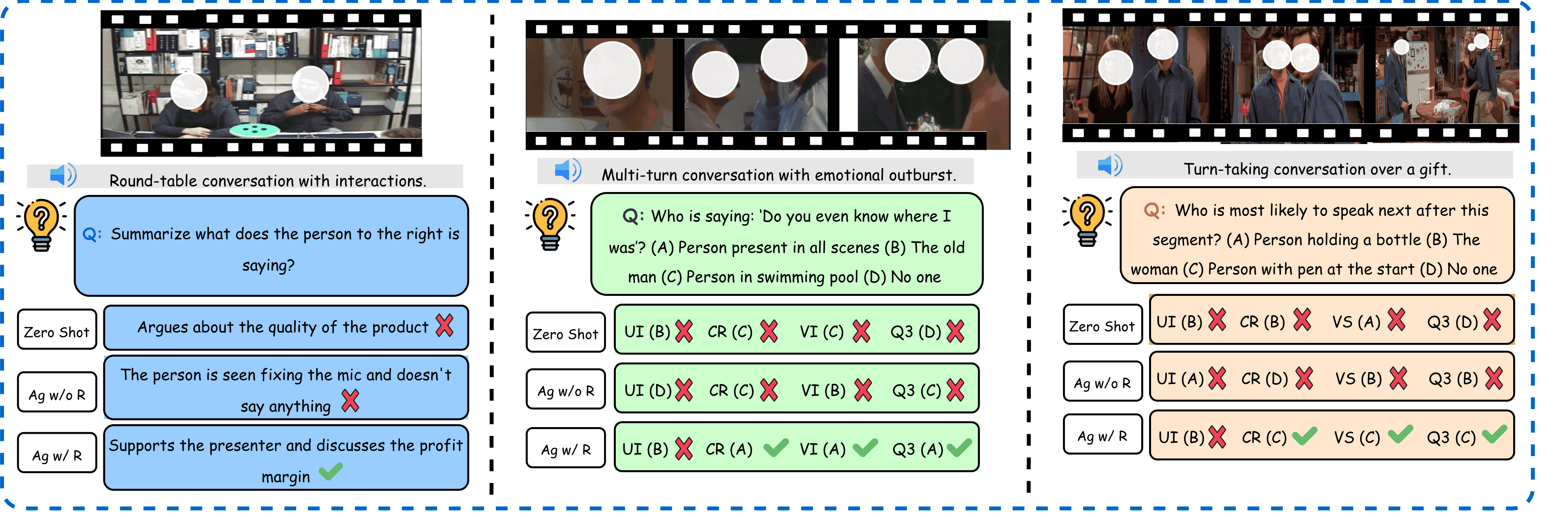}
\caption{\textbf{Qualitative results 1.} Comparison on multi-speaker reasoning tasks: Audio-Visual Dialogue Summarization (left), Speaker Association (middle), and Next Speaker Prediction (right).  UI: \textit{Unified-IO2}, CR: \textit{CREMA}, VS: \textit{VideoSALMONN}, VI: \textit{VITA}, Q2.5: \textit{Qwen2.5-Omni}, and  Q3: \textit{Qwen3-Omni} under Zero-Shot, Agentic w/o \oursolution, and Agentic w/ \oursolution modes. Results for AVDS is for Qwen3-Omni.}
\label{fig:qual-supp1}
\end{figure*}

\begin{figure*}[h]
\centering
\includegraphics[width=1\textwidth]{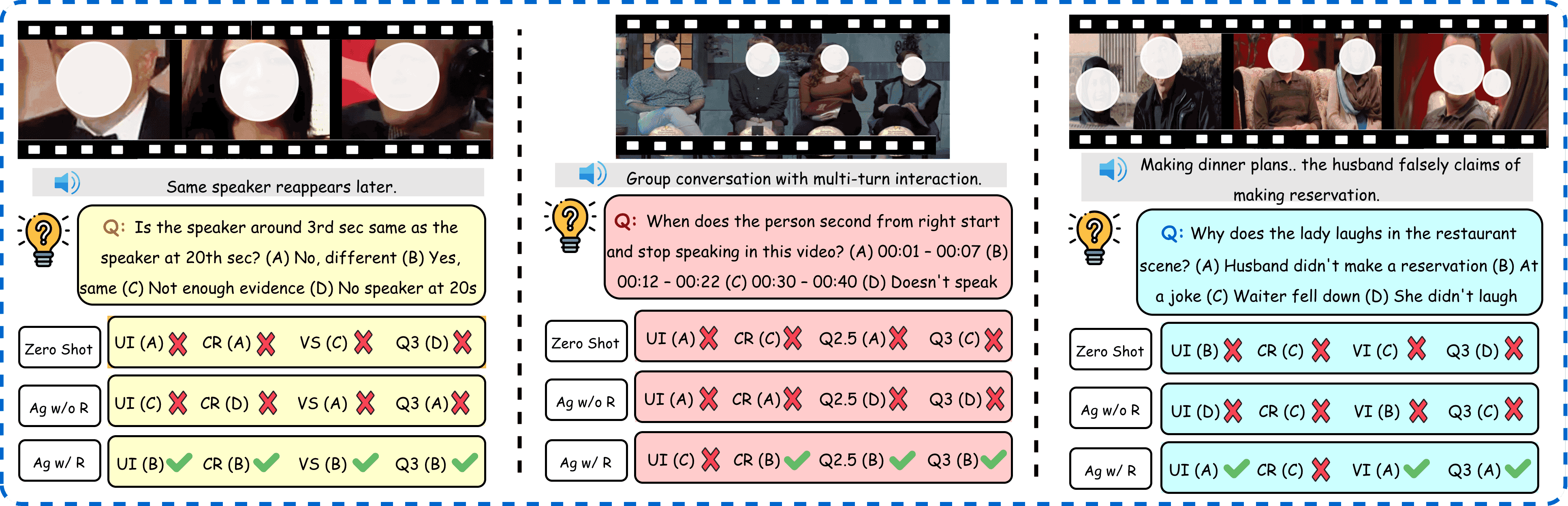}
\caption{\textbf{Qualitative results 2.} Comparison on multi-speaker reasoning tasks: Speaker Re-identification (left), Temporal Grounding (middle), and Cross Scene Narrative Linking (right).  UI: \textit{Unified-IO2}, CR: \textit{CREMA}, VS: \textit{VideoSALMONN}, VI: \textit{VITA}, Q2.5: \textit{Qwen2.5-Omni}, and  Q3: \textit{Qwen3-Omni} under Zero-Shot, Agentic w/o \oursolution, and Agentic w/ \oursolution modes.}
\label{fig:qual-supp2}
\end{figure*}

Figures~\ref{fig:qual-supp1} and \ref{fig:qual-supp2} illustrate qualitative comparisons across all six \ourbenchmark tasks under Zero-Shot, Agentic w/o \oursolution, and Agentic w/ \oursolution modes. We observe consistent improvements in multimodal grounding, speaker attribution, and temporal consistency when using \oursolution.

Fig. \ref{fig:qual-supp1} (AVDS, AVSA, NSP). Zero-shot models frequently rely on textual priors and ignore speaker cues, leading to incorrect summaries, mismatched utterance–speaker assignments, and poor turn-taking prediction. Agentic inference without \oursolution improves tool usage but remains unstable. In contrast, \oursolution enables accurate identification of the correct speaker in AVDS, reliable association of utterances in AVSA, and context-aware prediction of the next speaker in NSP by enforcing structured reasoning and perceptual alignment.

Fig. \ref{fig:qual-supp2} (SRID, STG, CSNL). For identity reasoning (SRID), non-\oursolution agents confuse visually similar individuals, while \oursolution reliably matches speakers across scenes. In STG, \oursolution reduces temporal drift and accurately localizes when a specific person starts or stops speaking under heavy overlap. In CSNL, \oursolution correctly links causally dependent events across disjoint scenes, avoiding shallow pattern matching. Overall, \oursolution yields coherent, grounded, and stable multi-speaker reasoning, complementing the quantitative gains reported in the main paper.

\section{User Study}
\label{user study}


\noindent\textbf{Sample Curation Validity.}
Each dataset sample is manually reviewed by human annotators to ensure accuracy and clarity. Raters watch the full clip, verify transcripts, speaker identities, and temporal spans, and check that the question unambiguously matches the underlying audio–visual evidence. Samples with unclear boundaries, mismatched associations, or ambiguous narratives are corrected or discarded, ensuring that all items used for evaluation are high-quality and reliable.

\noindent\textbf{Human Performance Estimation.}
To establish an upper bound on task difficulty, human raters also answer the evaluation questions themselves under the same conditions as the model. Annotators select answers for multiple-choice tasks or mark temporal segments for grounding tasks, providing a ceiling for achievable performance and helping distinguish true model errors from inherently ambiguous cases.

\section{\oursolution Pseudocode}
\label{raft algo}

Pseudocode. \ref{lst:raft} summarizes the \oursolution training procedure. The model first aligns its step-by-step reasoning to human supervision through the structured reasoning loss. It then samples multiple candidate responses and scores them with a perceptual reward to perform RRO, producing stable, reward-weighted updates. A temporal grounding regularizer enforces cross-modal synchrony across audio, visual, and textual streams. Finally, only the SRA adapter parameters are updated using the combined \oursolution objective, enabling efficient and well-grounded multimodal reasoning.

\renewcommand{\lstlistingname}{}
\begin{figure*}[t]
\vspace{-0.5em}
\begin{lstlisting}[language=Python,
caption={PyTorch-style pseudocode for RAFT (Reasoning, Acting, and Feedback Training)},
label={lst:raft}]
def RAFT_training():
    ############################################################
    # Reasoning, Acting, and Feedback Training (RAFT)
    ############################################################

    theta = initialize_model()
    insert_SRA_adapters(theta)

    while not converged:

        batch = sample_minibatch(D)

        ########################################################
        # 1. Structured reasoning alignment
        ########################################################
        y_k = sample_reasoning_steps(pi_theta, batch)
        L_align = -sum(log_prob(pi_theta, y_k))

        ########################################################
        # 2. Reflective Reward Optimization (RRO)
        ########################################################
        rewards = []
        for x in batch:
            y_samples = sample_K_outputs(pi_theta, x, K)
            rewards.append([R(x, y) for y in y_samples])

        r_bar = mean(rewards)
        weights = softmax(beta * (rewards - r_bar))
        J_RRO = sum(weights * log_prob(pi_theta, y_samples))

        ########################################################
        # 3. Temporal grounding regularization
        ########################################################
        f_a, f_v, f_s, f_r = extract_embeddings(batch)
        L_temp = sum(
            l2(f_a[t] - f_v[t]) +
            gamma * l2(f_s[t] - f_r[t])
            for t in time_steps
        )

        ########################################################
        # 4. RAFT objective and update
        ########################################################
        L_RAFT = L_align + alpha * L_temp - beta * J_RRO
        theta = gradient_step(theta, L_RAFT)

    return theta
\end{lstlisting}
\vspace{-0.5em}
\end{figure*}

\section{Evaluations Details}
\label{eval details}

\subsection{Prompt Templates Across Evaluation Modes}
We provide complete prompt templates for all three evaluation modes—zero-shot, guided, and agentic—including examples of how perception tools (ASR, diarization, face recognition, and AV sync) are incorporated. Zero-shot templates appear in the main paper; below we describe the guided and agentic settings. In the guided mode, all external tools (Whisper, PyAnnote, InsightFace, SyncNet) are executed offline and their outputs are inserted into the prompt as structured metadata, which the model must rely on without invoking tools itself. In the agentic mode, the model is instead given access to the full toolset and must autonomously decide when and how to call tools, integrate their outputs, and perform multi-step reasoning to solve each task.

\noindent\textbf{Guided Mode Prompt.} In the guided mode, the model operates purely as a reasoning layer over precomputed structured information (Tab. \ref{tab:guided-prompt-supp}). The template below is instantiated separately for each of the six \ourbenchmark tasks.

\begin{table*}[t]
\centering
\fcolorbox{black}{blue!5}{%
  \begin{minipage}{0.90\textwidth}
  \small
  \textbf{Guided-mode prompt template}\\[0.5em]
  You are given a video clip along with structured information extracted using external audio-visual tools. All processing has already been completed. Use only the information shown below to solve the task.\\[0.25em]
  \textbf{Video metadata:} short description, dataset source, and duration.\\
  \textbf{ASR transcript:} transcript text with timestamps.\\
  \textbf{Speaker diarization:} list of speech segments with start time, end time, and speaker identifiers.\\
  \textbf{Face tracks:} list of track identifiers, their visible time spans, and bounding-box intervals.\\
  \textbf{Audio-visual alignment:} optional synchronization offsets or scores.\\[0.25em]
  \textbf{Task description:} one of the six \ourbenchmark tasks (speaker temporal grounding, audio-visual dialogue summarization, speaker association, next speaker prediction, speaker re-identification, or cross-scene narrative linking).\\[0.25em]
  \textbf{Instructions:}\\
  (1) Read the task description carefully and determine what must be predicted.\\
  (2) Use transcript segments, diarization labels, face-track identifiers, and alignment information as explicit evidence.\\
  (3) Do not infer speakers, timestamps, or events that are not present in the structured fields.\\
  (4) When reasoning about speakers or faces, always refer back to the provided identifiers.\\
  (5) Produce a concise final answer and a short justification that cites the relevant segments or identifiers.
  \end{minipage}
}
\caption{Guided-mode prompt template. All external tools are executed before prompting, and their outputs are injected as structured text.}
\label{tab:guided-prompt-supp}
\end{table*}


\begin{table*}[t]
\centering
\fcolorbox{black}{blue!5}{%
  \begin{minipage}{0.90\textwidth}
  \small
  \textbf{Autonomous (agentic) prompt template}\\[0.5em]
  You are an audio-visual reasoning agent. You can decide when and how to use the following tools in order to solve the task:\\[0.25em]
  \textbf{Whisper:} transcribes the audio into text.\\
  \textbf{PyAnnote:} produces speaker diarization with time-stamped segments.\\
  \textbf{InsightFace:} provides face tracks and identity features for visible people.\\
  \textbf{SyncNet:} estimates audio-visual synchronization between speech and faces.\\[0.25em]
  \textbf{Task description:} one of the six \ourbenchmark tasks (speaker temporal grounding, audio-visual dialogue summarization, speaker association, next speaker prediction, speaker re-identification, or cross-scene narrative linking).\\[0.25em]
  \textbf{Instructions:}\\
  (1) First restate the task in your own words and outline what information you need.\\
  (2) Decide which tools are necessary to obtain that information and why; avoid unnecessary tool calls.\\
  (3) Invoke tools sequentially, update your plan after each result, and decide whether additional calls are required.\\
  (4) Treat tool outputs as ground-truth metadata and base your reasoning strictly on these results.\\
  (5) Do not hallucinate speakers, timestamps, or events that are not supported by tool outputs or the video description.\\
  (6) Once you have gathered sufficient evidence, produce a final answer along with a brief justification that explicitly cites the tool results and time segments you used.\\[0.25em]
  Your response should therefore contain: (a) a short plan, (b) references to the tools you chose to use and their returned outputs, and (c) a final answer with clear, evidence-based reasoning.
  \end{minipage}
}
\caption{Autonomous-mode (agentic) prompt template. The model independently determines which tools to invoke and integrates their outputs.}
\label{tab:agentic-prompt-supp}
\end{table*}

\noindent\textbf{Agentic Mode Prompt.} In the autonomous mode, the model acts as a multimodal agent. It must plan, call tools selectively, incorporate returned evidence, and synthesize a final answer. The tool calling details are reported in Tab. \ref{tab:tool-api-snippets} The same template is used across all \ourbenchmark tasks. The prompt used is explained in Tab. \ref{tab:agentic-prompt-supp}.

\subsection{More Details on LLM-based Choice Extraction}

\noindent\textbf{Choice extraction strategy.}
\label{choice extraction strategy appendix}
We adopt a two-stage procedure to robustly extract discrete choices from free-form AVLLM predictions. Although humans can easily infer the intended choice, rule-based matching is often brittle when faced with stylistic variation or incomplete responses. To ensure consistency across AVLLMs with diverse instruction-following abilities, we standardize the evaluation pipeline as follows:

\noindent\textbf{\textit{Step 1. Prediction matching:}} 
We first apply a lightweight heuristic matching strategy to directly detect the choice label (e.g., `A', `B', `C', `D') from the model’s output. If a valid label is found, it is used as the final prediction. If no reliable match is extracted, we proceed to the LLM-based extraction step.

\noindent\textbf{\textit{Step 2. GPT-4 processing:}}
Following prior benchmarks such as \cite{mmbench}, GPT-4 serves as a dependable choice extractor. When Step~1 fails, we provide GPT-4 with the question, the list of answer choices, and the model's free-form response, and instruct it to align the response with the most semantically similar option. If no option aligns, GPT-4 outputs ``No match found''. We additionally employ the CircularEval protocol \cite{mmbench} to ensure rigorous evaluation and to highlight performance differences among AVLLMs.

\noindent\textbf{Response matching.}
We treat an option as selected whenever it is referenced through its isolated label (e.g., `A') or standard labeled formats such as `A) \texttt{<response>}', `A. \texttt{<response>}', `A, \texttt{<response>}', or `(A) \texttt{<response>}' provided the \texttt{<response>} segment does not contain other option labels.

\noindent\textbf{Where does heuristic matching fail?}
Heuristic matching commonly fails in two situations:  
(i) when the AVLLM does not commit to an answer and instead asks for clarification (e.g., ``Apologies, could you clarify...?''), and  
(ii) when the model outputs multiple option labels simultaneously.  
In such cases, we defer to GPT-4 for choice extraction, as shown below.

\section{More Related Works}
\label{More Related Works}

\subsection{Preference Optimization Algorithms}

Early RLHF approaches \cite{ouyang2022training, bai2022training} optimized the RLHF objective using Proximal Policy Optimization (PPO) \cite{schulman2017proximal}. To improve memory efficiency and stability, later works removed the critic, introducing REINFORCE Leave-One-Out (RLOO) \cite{ahmadian2024back} and Group Relative Policy Optimization (GRPO) \cite{shao2402deepseekmath}, which were used in aligning models such as LLaMA3-Nemotron-70B-Instruct and DeepSeek-R1 \cite{wang2024helpsteer2preference, guo2025deepseek}. Beyond online RL, Direct Preference Optimization (DPO) \cite{rafailov2024direct} enables offline alignment by directly optimizing the policy via an implicit reward model, inspiring numerous variants including IPO, SimPO, KTO, DNO, and related extensions \cite{mitchell2023note, azar2024general, meng2024simpo, ethayarajh2024kto, rosset2024direct}. Recent methods further explore iterative and supervised alignment \cite{chenself} or emphasize backward KL and offline training \cite{adler2024nemotron}. Building on this line of work, we generalize RPO to multiple design choices, positioning it as a unifying framework bridging online and offline preference optimization.

\subsection{Multi-modal Learning}
Multimodal learning, beyond single-modality approaches \cite{vdesirr, maw}, has seen rapid progress in cross-modal generation \cite{adverb, melfusion, foleygen, tang2024codi, magnet}, audio-visual representation learning \cite{listentopixel, audvisum, gao2024audio, sudarsanam2025representation, egoadapt}, multimodal large language models \cite{aurelia, meerkat, avtrustbench, aura}, and cross-modal integration \cite{apollo, intentometer, vlmnav}. Recent work leverages visual and linguistic context to synthesize coherent audio \cite{adverb, melfusion}, while studies on active audio-visual separation and embodied agents highlight the role of motion and egocentric perception in learning robust representations. These ideas extend naturally to audio-visual LLMs \cite{vlmnav, avnav}, where perceptually grounded models interact with dynamic environments, underscoring the importance of alignment and embodied perception for versatile multimodal systems \cite{apollo}.

\begin{tcolorbox}[
    float,
    width=\columnwidth,
    colback=white,
    colframe=cyan!25!white,
    colbacktitle=cyan!10!white,
    coltitle=black,
    before skip=0pt,
    after skip=0pt,
    top=2pt, bottom=2pt, left=2pt, right=2pt,
    boxrule=0.4pt,
    title=\textcolor{black}{Choice extraction prompt for GPT-4}
]
\small
Can you help me match an answer with a set of options for a single-correct-answer question?  
I will provide a question, a set of options, and a model-generated response.  
Your task is to map the response to the most similar option. Output exactly one uppercase letter from \{A, B, C, D, E\}.  
If no option matches, respond with ``No match found''.  
Please avoid subjectivity and do not use external knowledge.

\textbf{Example 1:}\\
\textit{Question:} What color is the man's shirt who is sitting left of the object making this sound?\\
\textit{Options:} A. Green \quad B. Red \quad C. Yellow \quad D. Black\\
\textit{Answer:} The person sitting next to the record player is wearing a black shirt.\\
\textit{Your output:} D

\textbf{Example 2:}\\
\textit{Question:} What does the audio-visual event constitute?\\
\textit{Options:} A. A dog barking at a cat \quad 
B. A dog barking on being hit by a stick \quad 
C. The dog is hungry \quad 
D. The dog is chasing another dog\\
\textit{Answer:} It is a wolf.\\
\textit{Your output:} No match found
\end{tcolorbox}




\begin{table*}[h]
\centering
\begin{tcolorbox}[colback=white, colframe=SummBlue, boxrule=0.6pt, arc=3pt] 
\resizebox{\linewidth}{!}{
\begin{tabular}{p{0.97\linewidth}}
\rowcolor{SummBlue}
\textbf{Prompt Variants for Audio-Visual Dialogue Summarization} \\
\hdashline
1. Summarize what the person in the \textless descriptor\textgreater{} says between \textless time\_token\textgreater{} and \textless time\_token\textgreater{}. \\
2. What is the main idea expressed by the speaker wearing a red shirt at \textless time\_token\textgreater{}? Please summarize \\
3. Briefly summarize the key point the woman on the left conveys during \textless time\_token\textgreater{}. \\
4. What is the speaker in the blue jacket trying to communicate at \textless time\_token\textgreater{}? \\
5. Summarize the statement made by the man in the center into one sentence. \\
6. What message is the person standing on the right conveying at \textless time\_token\textgreater{}? \\
7. Describe what the highlighted individual talks about during the segment starting at \textless time\_token\textgreater{}. \\
8. Provide a brief summary of the response given by the seated person at \textless time\_token\textgreater{}. \\
9. Rephrase the speaker’s comment between \textless time\_token\textgreater{} and \textless time\_token\textgreater{} concisely. \\
10. What conclusion does the person in the black hoodie present during this segment? \\
11. In a few words, describe what the speaker with glasses emphasizes at \textless time\_token\textgreater{}. \\
12. What information does the person in \textless outfit descriptor\textgreater{} share at \textless time\_token\textgreater{}? \\
13. What does the dialogue turn from the woman on the right mainly focus on at \textless time\_token\textgreater{}? \\
14. Summarize the viewpoint expressed by the man in the gray shirt in this part. \\
15. What does the person near the doorway explain during the segment starting at \textless time\_token\textgreater{}? \\
16. Which topic does the speaker in the red dress address between \textless time\_token\textgreater{} and \textless time\_token\textgreater{}? \\
17. Summarize the line spoken by the person on the left couch at \textless time\_token\textgreater{}. \\
18. What is the essence of the statement made by the speaker standing at the table? \\
19. Give a short paraphrase of what the person in the blue sweater says at \textless time\_token\textgreater{}. \\
20. Summarize the main point communicated by the speaker facing the camera. \\
\end{tabular}}
\end{tcolorbox}
\caption{Prompt variants for the \textbf{Audio-Visual Dialogue Summarization} task.}
\label{tab: question avds}
\end{table*}

\begin{table*}[h]
\centering
\begin{tcolorbox}[colback=white, colframe=AssocGreen, boxrule=0.6pt, arc=3pt]
\begin{tabular}{p{0.97\linewidth}}
\rowcolor{AssocGreen}
\textbf{Prompt Variants for Audio-Visual Speaker Association} \\
\hdashline
1. Who is speaking during the audio segment at \textless time\_token\textgreater{}? \\
2. Match the utterance at \textless time\_token\textgreater{} to the correct person in the scene. \\
3. Which individual is producing the speech at \textless time\_token\textgreater{}? \\
4. Identify who is talking using lip movement and voice cues at \textless time\_token\textgreater{}. \\
5. Who is talking while others remain silent at \textless time\_token\textgreater{}? \\
6. Which person corresponds to the audio clip starting at \textless time\_token\textgreater{}? \\
7. Who is the active speaker when the man in the red shirt moves his lips at \textless time\_token\textgreater{}? \\
8. Whose voice do we hear when the woman seated on the left is shown at \textless time\_token\textgreater{}? \\
9. Based on audio-visual cues, who is speaking while the man in the blue jacket appears at \textless time\_token\textgreater{}? \\
10. Which on-screen person is talking during the segment at \textless time\_token\textgreater{}? \\
11. Whose lip motion aligns with the spoken sentence at \textless time\_token\textgreater{}? \\
12. Identify the speaker when the right-side participant is visible at \textless time\_token\textgreater{}. \\
13. Which speaker’s voice corresponds to the utterance at \textless time\_token\textgreater{}? \\
14. Which person produces the spoken line heard at \textless time\_token\textgreater{}? \\
15. Who is responsible for the highlighted phrase at \textless time\_token\textgreater{}? \\
16. Who is delivering the dialogue while the person in the black jacket is centered at \textless time\_token\textgreater{}? \\
17. Which person should be attributed as the speaker of the sentence aligned with the lip motion at \textless time\_token\textgreater{}? \\
18. Whose mouth movement matches the audio when the left side of the table is shown at \textless time\_token\textgreater{}? \\
19. Who is the speaker when the person in the white shirt appears at \textless time\_token\textgreater{}? \\
20. Who produces the spoken line associated with the audio segment at \textless time\_token\textgreater{}? \\
\end{tabular}
\end{tcolorbox}
\caption{Prompt variants for the \textbf{Audio-Visual Speaker Association} task.}
\label{tab: question avsa}
\end{table*}

\begin{table*}[t]
\centering
\begin{tcolorbox}[colback=white, colframe=NextOrange, boxrule=0.6pt, arc=3pt]

\begin{tabular}{p{0.97\linewidth}}
\rowcolor{NextOrange}
\textbf{Prompt Variants for Next Speaker Prediction} \\
\hdashline
1. Based on the interaction up to \textless time\_token\textgreater{}, who is most likely to speak next? \\
2. Who seems prepared to reply following the segment ending at \textless time\_token\textgreater{}? \\
3. Which individual is most likely to take the next turn in the conversation? \\
4. Predict the next speaker among the on-screen participants. \\
5. Who appears ready to answer the question asked at \textless time\_token\textgreater{}? \\
6. Which person is positioned to speak next, given their posture and gaze? \\
7. Whose body language suggests they are about to answer after \textless time\_token\textgreater{}? \\
8. Considering the conversation flow, who is expected to continue the dialogue? \\
9. Who on the left side of the frame seems ready to speak next? \\
10. Which person on the right side of the table will likely speak after the current turn? \\
11. Who follows up the conversation after the speaker in the red shirt finishes at \textless time\_token\textgreater{}? \\
12. Using gaze direction and facial expressions at \textless time\_token\textgreater{}, who is likely to speak next? \\
13. Which person will likely contribute the next line following \textless time\_token\textgreater{}? \\
14. Which participant seated on the couch is expected to speak next? \\
15. From the pattern of turn-taking up to \textless time\_token\textgreater{}, who takes the next turn? \\
16. Who seems about to interject when the camera shows the group at \textless time\_token\textgreater{}? \\
17. Which character is cueing up the next utterance, for example by leaning forward or opening their mouth? \\
18. Whose gestures indicate that they are preparing to speak next? \\
19. Who resumes the conversation after the short pause at \textless time\_token\textgreater{}? \\
20. Who logically continues the dialogue when the question is directed towards the person in the blue sweater at \textless time\_token\textgreater{}? \\
\end{tabular}
\end{tcolorbox}
\caption{Prompt variants for the \textbf{Next Speaker Prediction} task.}
\label{tab: question nsp}
\end{table*}

\begin{table*}[t]
\centering
\begin{tcolorbox}[colback=white, colframe=TempPurple, boxrule=0.6pt, arc=3pt]

\begin{tabular}{p{0.97\linewidth}}
\rowcolor{TempPurple}
\textbf{Prompt Variants for Speaker Temporal Grounding} \\
\hdashline
1. At what \textless time\_token\textgreater{} does the person in \textless descriptor\textgreater{} begin speaking? \\
2. When does the woman in the red dress start talking in the video? \\
3. Identify the \textless time\_token\textgreater{} at which the man on the left first begins to speak. \\
4. Locate the moment the person in the blue shirt starts speaking. \\
5. At which \textless time\_token\textgreater{} does their speech initiation occur? \\
6. Find the starting time of the speaker’s voice for the person near the doorway. \\
7. When is the first audible word from the person sitting on the right side of the table? \\
8. Mark the \textless time\_token\textgreater{} at which the person in the black hoodie begins their utterance. \\
9. What is the earliest \textless time\_token\textgreater{} at which this person starts speaking? \\
10. Between which \textless time\_token\textgreater{} values does the speaker’s utterance begin? \\
11. Give the \textless time\_token\textgreater{} where the woman in the center first speaks. \\
12. When does the speech associated with the man in the gray sweater start? \\
13. Identify the first frame in time (as \textless time\_token\textgreater{}) when the speaker on the left starts talking. \\
14. At what \textless time\_token\textgreater{} does the dialogue contribution of the person in the white shirt begin? \\
15. At what time does this speaker enter the conversation for the first time? \\
16. When is their first vocalization heard after they appear on screen? \\
17. What exact \textless time\_token\textgreater{} corresponds to the onset of the speaker’s voice? \\
18. Find the \textless time\_token\textgreater{} where this speaker’s sentence begins in the timeline. \\
19. At which \textless time\_token\textgreater{} does the person standing at the counter start speaking? \\
20. Determine the onset \textless time\_token\textgreater{} of the utterance produced by the speaker in \textless descriptor\textgreater{}. \\
\end{tabular}
\end{tcolorbox}
\caption{Prompt variants for the \textbf{Speaker Temporal Grounding} task.}
\label{tab: question stg}
\end{table*}

\begin{table*}[t]
\centering
\begin{tcolorbox}[colback=white, colframe=ReIDYellow, boxrule=0.6pt, arc=3pt]

\begin{tabular}{p{0.97\linewidth}}
\rowcolor{ReIDYellow}
\textbf{Prompt Variants for Speaker Re-identification} \\
\hdashline
1. Is the speaker at \textless time\_token\textgreater{} the same person as the speaker at \textless time\_token\textgreater{}? \\
2. Does the voice in the segment at \textless time\_token\textgreater{} match the voice at \textless time\_token\textgreater{}? \\
3. Are the speech segments at the two \textless time\_token\textgreater{} values produced by the same individual? \\
4. Is the person wearing a red shirt at \textless time\_token\textgreater{} the same as the speaker at \textless time\_token\textgreater{}? \\
5. Compare the speaker at \textless time\_token\textgreater{} with the speaker at \textless time\_token\textgreater{}: are they the same person? \\
6. Do the face and voice at \textless time\_token\textgreater{} correspond to the same identity as at \textless time\_token\textgreater{}? \\
7. Are the appearances of the man on the left at the two \textless time\_token\textgreater{} positions from the same person? \\
8. Is the person in the blue jacket at \textless time\_token\textgreater{} the same speaker who talks at \textless time\_token\textgreater{}? \\
9. Do the vocal patterns and facial cues at \textless time\_token\textgreater{} and \textless time\_token\textgreater{} indicate a single speaker identity? \\
10. Is the speaker near the doorway at \textless time\_token\textgreater{} the same as the speaker at \textless time\_token\textgreater{}? \\
11. Does the speaker at \textless time\_token\textgreater{} match the person in the striped shirt speaking at \textless time\_token\textgreater{}? \\
12. Are the speakers across the segments at \textless time\_token\textgreater{} and \textless time\_token\textgreater{} the same individual? \\
13. Is the woman on the right speaking at \textless time\_token\textgreater{} the same woman speaking at \textless time\_token\textgreater{}? \\
14. Are the vocal characteristics of the person in the black hoodie at \textless time\_token\textgreater{} consistent with those at \textless time\_token\textgreater{}? \\
15. Does the person speaking at the table at \textless time\_token\textgreater{} correspond to the same identity speaking on the couch at \textless time\_token\textgreater{}? \\
16. Is the speaker shown in close-up at \textless time\_token\textgreater{} the same as the one talking in the wide shot at \textless time\_token\textgreater{}? \\
17. Does the voice of the person in the red dress at \textless time\_token\textgreater{} match the voice at \textless time\_token\textgreater{}? \\
18. Do the face and audio cues of the man on the left at \textless time\_token\textgreater{} indicate the same speaker identity as at \textless time\_token\textgreater{}? \\
19. Does the person shown near the window at \textless time\_token\textgreater{} match the speaker filmed near the table at \textless time\_token\textgreater{}? \\
20. Are the dialogue segments at \textless time\_token\textgreater{} and \textless time\_token\textgreater{} delivered by the same speaker? \\
\end{tabular}
\end{tcolorbox}
\caption{Prompt variants for the \textbf{Speaker Re-identification} task.}
\label{tab: question sri}
\end{table*}

\begin{table*}[t]
\centering
\begin{tcolorbox}[colback=white, colframe=NarrTeal, boxrule=0.6pt, arc=3pt]

\begin{tabular}{p{0.97\linewidth}}
\rowcolor{NarrTeal}
\textbf{Prompt Variants for Cross-scene Narrative Linking} \\
\hdashline
1. How does the event at \textless time\_token\textgreater{} connect to the reaction of the person in the red sweater at \textless time\_token\textgreater{}? \\
2. What detail shown at \textless time\_token\textgreater{} explains why the woman on the left reacts at \textless time\_token\textgreater{}? \\
3. Why does the man in the blue shirt react the way he does at \textless time\_token\textgreater{}, given what happened at \textless time\_token\textgreater{}? \\
4. Which event at \textless time\_token\textgreater{} provides context for the final scene at \textless time\_token\textgreater{}? \\
5. What narrative link exists between the segment at \textless time\_token\textgreater{} and the segment at \textless time\_token\textgreater{}? \\
6. How does the phone call or object mention at \textless time\_token\textgreater{} relate to the reaction of the person standing on the right at \textless time\_token\textgreater{}? \\
7. What realization does the woman in the black jacket have at \textless time\_token\textgreater{} based on something shown at \textless time\_token\textgreater{}? \\
8. Explain what triggers the behavior of the person sitting on the couch at \textless time\_token\textgreater{}, using clues from \textless time\_token\textgreater{}. \\
9. Which event witnessed by the man on the far left at \textless time\_token\textgreater{} leads to his action at \textless time\_token\textgreater{}? \\
10. How does the interaction at \textless time\_token\textgreater{} influence the character’s response at \textless time\_token\textgreater{}? \\
11. What causal link connects the scene at \textless time\_token\textgreater{} with the outcome at \textless time\_token\textgreater{}? \\
12. Which visual clue shown at \textless time\_token\textgreater{} helps explain the reaction of the woman in the red dress at \textless time\_token\textgreater{}? \\
13. How does the introduction of the \textless object\textgreater{} at \textless time\_token\textgreater{} shape the character’s interpretation at \textless time\_token\textgreater{}? \\
14. Which detail noticed by the man on the right side of the frame at \textless time\_token\textgreater{} sets up his reaction at \textless time\_token\textgreater{}? \\
15. How does the sequence at \textless time\_token\textgreater{} prepare the narrative moment occurring at \textless time\_token\textgreater{}? \\
16. What information revealed at \textless time\_token\textgreater{} does the woman in the blue sweater realize at \textless time\_token\textgreater{}? \\
17. What continuity links the segment occurring at \textless time\_token\textgreater{} with the one at \textless time\_token\textgreater{}? \\
18. How do the events at \textless time\_token\textgreater{} and \textless time\_token\textgreater{} form a complete narrative arc? \\
19. What observation made by the person in the striped shirt at \textless time\_token\textgreater{} is recalled at \textless time\_token\textgreater{}? \\
20. What chain of events starting from the scene at \textless time\_token\textgreater{} leads to the reaction of the person near the doorway at \textless time\_token\textgreater{}? \\
\end{tabular}
\end{tcolorbox}
\caption{Prompt variants for the \textbf{Cross-scene Narrative Linking} task.}
\label{tab: question csnl}
\end{table*}

\clearpage
\raggedbottom
\clearpage



{
    \small
    \bibliographystyle{ieeenat_fullname}
    \bibliography{main}
}

\end{document}

%% file: apple_preamble.tex
\usepackage{amsmath}
\usepackage{enumerate}
\usepackage{algorithm}
\usepackage{algpseudocode}
\usepackage{amsfonts}
\usepackage{amsthm}
\usepackage{cleveref}
\usepackage{diagbox}
\usepackage{colortbl}
\usepackage{amssymb}
\usepackage{xspace}
\usepackage{wrapfig}
\usepackage{adjustbox}
\usepackage{tabularx}
\usepackage{booktabs}
\usepackage{mathtools}
\usepackage{tikz}
\usepackage{enumitem}
\usepackage{silence}
\usepackage{dsfont}
\usepackage[table]{xcolor}
\usepackage[dvipsnames]{xcolor}
\usepackage{multirow}
\usepackage{makecell}
\usepackage{xfakebold}
\input{math_commands}

\definecolor{textgray}{HTML}{6E6E73}
\usetikzlibrary{positioning, calc}
\usetikzlibrary{decorations.pathmorphing}

\makeatletter
\patchcmd{\wrong@fontshape}{\@gobbletwo}{}{}{}
\makeatother
\numberwithin{equation}{section}
\makeatletter
\AtBeginDocument{
  \urlstyle{sf}
  
}
\makeatother

\definecolor{light}{RGB}{125, 125, 125}
\crefname{tcb@cnt@pbox}{code}{code}
\Crefname{tcb@cnt@pbox}{Code}{Code}
\crefname{assumption}{assumption}{assumption}
\Crefname{assumption}{Assumption}{Assumptions}

\newtcolorbox[auto counter]{pbox}[2][]{
  colback=white,
  title=Code~\thetcbcounter: #2,
  #1,fonttitle=\sffamily,
  fontupper=\sffamily,
  arc=2pt,
  colframe=bgcolor,
  coltitle=fgcolor,
  colbacktitle=bgcolor,
  toptitle=0.25cm,
  bottomtitle=0.125cm
}

\makeatletter
\newcommand\applefootnote[1]{%
  \begingroup
  \renewcommand\thefootnote{}%
  \renewcommand\@makefntext[1]{\noindent##1}%
  \footnote{#1}%
  \addtocounter{footnote}{-1}%
  \endgroup
}
\makeatother

\definecolor{cverbbg}{gray}{0.90}

\newcommand{\eq}{\texttt}

%% file: math_commands.tex
\usepackage{amsmath,amsfonts,bm}

\def\eqref#1{equation~\ref{#1}}

\def\1{\bm{1}}

\DeclareMathAlphabet{\mathsfit}{\encodingdefault}{\sfdefault}{m}{sl}
\SetMathAlphabet{\mathsfit}{bold}{\encodingdefault}{\sfdefault}{bx}{n}

